# A review on development of eco-friendly filters in Nepal for use in cigarettes and masks and Air Pollution Analysis with Machine Learning and SHAP Interpretability

Bishwash Paneru[a,*], Biplov Paneru[b], Tanka Mukhiya[a], Khem Narayan Poudyal[a]

[a]Department of Applied Sciences and Chemical Engineering, Institute of Engineering, Pulchowk Campus, Tribhuvan University, Lalitpur 44700, Nepal

[b]Department of Electronics, Communication and Information Engineering, Nepal Engineering College, Pokhara University, Bhaktapur, Nepal

*Corresponding author: Bishwash Paneru
Email: 076bch015.bishwash@pcampus.edu.np

**Abstract:**
In Nepal, air pollution is a serious public health concern, especially in cities like Kathmandu where particulate matter (PM2.5 and PM10) has a major influence on respiratory health and air quality. The Air Quality Index (AQI) is predicted in this work using a Random Forest Regressor, and the model's predictions are interpreted using SHAP (SHapley Additive exPlanations) analysis. With the lowest Testing RMSE (0.23) and flawless $R^2$ scores (1.00), CatBoost performs better than other models, demonstrating its greater accuracy and generalization which is cross validated using a nested cross validation approach. NowCast Concentration and Raw Concentration are the most important elements influencing AQI values, according to SHAP research, which shows that the machine learning results are highly accurate. Their significance as major contributors to air pollution is highlighted by the fact that high values of these characteristics significantly raise the AQI. This study investigates the Hydrogen-Alpha (HA) biodegradable filter as a novel way to reduce the related health hazards. With removal efficiency of more than 98% for PM2.5 and 99.24% for PM10, the HA filter offers exceptional defense against dangerous airborne particles. These devices, which are biodegradable face masks and cigarette filters, address the environmental issues associated with traditional filters' non-biodegradable trash while also lowering exposure to air contaminants. This study demonstrates a sustainable strategy for addressing air pollution by fusing the excellent filtration capability of HA technology with data-driven insights from SHAP analysis. Significant ecological and health benefits can be obtained by combining predictive modeling with eco-friendly solutions, especially in areas like Nepal where air quality and waste management are serious problems. This study shows how machine learning and novel materials can be used to successfully address related environmental and public health issues.

## 1. Introduction

The little part at the end of a cigarette called a filter reduces the amount of dangerous chemicals that might enter the smoker's body and negatively impact their health. In 1931, the premium cigarette filter brand Parliament (Benson and Hedges) was introduced to the market (Pauly et al., 2009). The actual cigarette was a premium-filtered cigarette. In 1936, Viceroy, the first filtered cigarette with a cork tip, was invented (https://tobaccotactics.org). Around 70 mm in length, cigarettes were usually unfiltered at the time, and all brands were similar (Kozlowski et al., 2000).

In 1953, cellulose acetate fiber-based cork-tipped cigarettes were first made commercially available. In 1940, filters were promoted as a novelty product to attract female smokers (Browne CL, 1990). The most effective barrier against cigarette smoke in human health history was the "Micronite Filter," developed by Lorillard in 1954 (Borio G, 2009). Again, similar vents were added to filters made in the 1970s, which let air into the mainline smoke (O'Connor et al., 2008). The parameters that were causing the smoke components to undergo retention both with and without the addition of activated carbon were then identified by Swiss scientists in the 1973s. Today, cigarette filters can be seen every day and everywhere ("Facile Synthesis of Porous Carbon/Fe$_3$O$_4$ Composites Derived from Waste Cellulose Acetate by One-Step Carbothermal Method as a Recyclable Adsorbent for Dyes," 2020; Kumar Raja Vanapalli et al., 2023; Parveen et al., 2024).

By the 1980s, almost all cigarettes sold all over the world had filters (93% market share, non-filter cigarettes, 7% market share) (Pauly et al., 2009). Also in same year, studies had been conducted which showed that smokers of low-yield, ventilated-filter ('less-hazardous') cigarettes sometimes successfully defeated the purpose of the smoke-dilution holes by occluding the holes (Kozlowski et al., 1980). In 2003, carbon filters were developed for cigarette to provide potentially less exposure to the carcinogens of conventional cigarettes but it failed to get significant customers and was stopped to be used (Laugesen et al., 2006). For making cigarette filters safer, various steps have been put forwarded, anti-toxic flavoring agents can be added to filters like described in (Bellamah et al., 2008), studies for determination of components of tobacco used in cigarettes have been conducted (Leanderson et al., 1997; Qamar and Sultana, 2011; Yang, 1958; Cuzin et al., 1965; Shifflett et al., 2017; Mi et al., 2015), and standard filters using compounds like polyphenols to remove contaminants (Catel-Ferreira et al., 2015) and new filters are being developed which can be used even in cigarettes and research for improving filtering efficiency by making improvement in ventilation are also being conducted (Kozlowski et al., 2006). Cigarette compounds that are used by enzymes or other elements in the arsenic methylation process. Furthermore, cigarettes themselves contain a small quantity of arsenic; after consuming one cigarette, 0.25 µg of arsenic was detected (ATSDR). Because smoking and groundwater consumption of arsenic have a synergistic impact, residents of areas with high concentrations of arsenic are at risk for major health effects (Watts et al., 2018).

With 38% of men and 6% of women consuming other kinds of tobacco and 30% of men and 9% of women smoking cigarettes, Nepal has the highest rates of tobacco usage in South Asia (Nepal | Global Action to End Smoking, 2021). One in Nepal, tobacco use is becoming a bigger problem and a public health issue. Major problem with smoking cigarettes is that it can result in development of cancer (AL-Hashimi et al., 2023). It adds to morbidity and mortality and raises the risk of a number of infectious and non-communicable illnesses, including cancer, heart disease, chronic respiratory disorders, and tuberculosis. The Nepalese government was working on a ban on e-cigarettes as of May 2024, which would forbid their manufacture, importation, distribution, sale, public use, and promotion. Although heated tobacco products (HTPs) are not prohibited by law, they are not sold in the nation. Health warnings on devices or e-liquid packaging are not necessary as of 2022. Regulations limiting e-cigarette flavors do not exist.

Face masks have been life-savior devices in the modern world. Masks may typically shield the wearer from breathing in dangerous gasses, tiny particles (like pollens or allergies), and airborne microorganisms (like germs or viruses) (Pandit et al., 2021). Masks, respirators, face shields, or goggles have shown their ability to protect from respiratory infections (Adib Bin Rashid & Nazmir-Nur Showva, 2023). The purpose of face masks, such as medical masks and other filtered varieties like N95 and KN95 respirators, is to shield users from germs, viruses, and airborne particles (O'Dowd et al., 2020). While medical masks with higher-grade filters offer more strong protection by filtering tiny particles, medical masks, which are usually composed of many layers with synthetic fiber filters, are crucial for healthcare environments. During and during the COVID-19 epidemic, these masks have been essential for safeguarding public health in Nepal, particularly in crowded metropolitan areas (Singh et al., 2021). There are several varieties of face masks, each designed for a certain defensive purpose. Disposable surgical masks composed of layers of synthetic materials, such as polypropylene, are used extensively in healthcare; they shield others from the wearer's respiratory emissions and block bigger droplets (Shirvanimoghaddam et al., 2022). A greater level of filtration is offered by N95 respirators, which are often used in industrial and medical environments. They block at least 95% of airborne particles and give better protection against tiny particles. The general population frequently wears cloth masks, which come in a variety of textiles, for basic protection, particularly in situations when medical-grade masks might not be required (Ju et al., 2021).

Face masks are an important public health measure in Nepal, providing crucial defense against respiratory diseases and extreme air pollution. Urbanization trends have brought many challenges, one of the most acute of which is air pollution (Liu et al., 2022). Masks were crucial in stopping the spread of the COVID-19 virus during the pandemic, particularly in crowded cities, protecting susceptible groups including the elderly and people with weakened immune systems (Wimalawansa, 2020). In addition to tobacco usage, air pollution from automobiles, building dust, and industrial pollutants is a common occurrence in Nepal's cities, including as Kathmandu, placing locals at risk for cardiovascular and respiratory illnesses. By blocking hazardous particulate matter, masks—especially those with filters—help lessen the negative health effects of extended exposure to pollution (Carlsten et al., 2020). Furthermore, donning a mask has come to represent civic duty and a dedication to individual and public health. Face masks are now widely accessible and reasonably priced, making them an economical way to promote health and wellbeing in Nepal.
In Nepal and other developing countries, cigarette and mask filters have a big impact on the environment and public health.

Cigarette filters, made of cellulose acetate, a non-biodegradable substance, are primarily used to reduce the amount of nicotine and dangerous tar that is consumed (Everaert et al., 2023). However, they often give smokers the erroneous sense that they are safe, and they do not considerably reduce health hazards. With 38% of men and 6% of women consuming other kinds of tobacco and 30% of men and 9% of women smoking cigarettes, Nepal has the highest rates of tobacco usage in South Asia. In Nepal, tobacco use is becoming a bigger problem and a public health issue. It adds to morbidity and mortality and raises the risk of a number of infectious and non-communicable illnesses, including cancer, heart disease, chronic respiratory disorders, and tuberculosis (Using Behaviour Change Interventions to Decrease Tobacco Use in Nepal POLICY BRIEF Using Behaviour Change Interventions to Decrease Tobacco Use in Nepal, n.d.). Smoking and geographic differences had certain effects on the risk of liver cancer (Li et al., 2024). Obviously, most of the tobacco is used in cheap filters in Nepal which are accused to be less effective against carcinogenic compounds like nitrosamines present in cigarette's smoke.

An estimated 5.7 trillion cigarettes were consumed worldwide in 2016, which translates to 1.2 million tons of cigarette butt discarded annually. By 2025, this number is predicted to rise by more than 50% (Akanyeti et al., 2020). Worldwide, cigarette usage is quite high. With no practical purpose, used cigarettes are being discarded on the ground or in the trash (Singh et al., 2020). Similarly in case of Nepal due to insufficient waste management systems, discarded cigarette butts accumulate in rivers and towns, worsening pollution by releasing dangerous chemicals and microplastics into the environment (Shah et al., 2023). According to the research by Neupane et al. (2019) on locally accessible facemasks in Nepal, Cloth Masks' filtering effectiveness varied between 63% and 84%. It was discovered that the surgical mask has a 94% filtering efficacy. They came to the conclusion that the main difference in efficiency is the cloth facemask's bigger pore size and any contributing circumstances, such as stretching, drying, and washing. According to the study by (Tcharkhtchi et al., 2021), the maks described with their comfortness and reusability and presented in the Table 1 are the types of masks common in use in different parts of the world along with developing country like Nepal.

**Table 1** Different types of face masks in use.

| Mask Type | Comfort | Reusable |
|---|---|---|
| Cloth mask | High | Yes |
| Surgical face mask | High | No |
| Full-length face shield | Moderate | Yes |
| N95 | High | No (not suitable for washing) |
| P100 respirator | Moderate | Yes |
| Self-contained breathing apparatus | Moderate | Yes |
| Full face respirator | Low | Yes |

In a similar vein, mask waste has increased due to the frequent usage of masks, particularly during periods of severe pollution or health emergencies like COVID-19. Economic limitations cause many people in Nepal and comparable places to rely on less expensive, less effective masks, such as surgical masks or single-layer fabric, which give poor protection against fine particulate matter (PM2.5) and other pollutants, even though masks like N95 offer good filtration. Because single-use masks frequently wind up in landfills or are burned, releasing harmful substances into the atmosphere, improper disposal of these masks also adds to pollution.

A multifaceted strategy is needed to address these problems, which includes promoting biodegradable alternatives, improving trash disposal and recycling facilities, and increasing public awareness of the environmental effects of cigarette filters and mask debris (Bhattacharjee et al., 2022). Encouraging the development of biodegradable cigarette filters through laws might reduce environmental strain, while providing subsidies for high-quality, reasonably priced masks would enhance health results. One of the most prevalent pollutants in the environment is waste cigarette filters (WCF) (Yusuf Wibisono et al., 2024). Nepal and other areas might lower pollution and enhance public health in this way, but long-term change requires concerted work at the local and governmental levels (Acharya, 2024).

Despite many perceptions that filters are highly defensive against toxic compounds of smoke like nitrosamines research has shown that cigarette filters do not offer much health benefit and filtered cigarettes are not less harmful than unfiltered cigarettes, whether for smokers or passive smokers (https://tobaccotactics.org). Potential solutions have been explored to make toxic compounds that emit during filtration to be removed. Thus far, the main approaches have focused on treating filters contaminated with toxicants as described by ISO mainstream smoke that yields of 43 toxicants that were measured from cigarettes containing treated tobaccos; lower yields of tar, nicotine, carbon monoxide (16–20%), acrylonitrile, ammonia, aromatic amines, pyridine, quinolene and hydrogen cyanide (33–51%), tobacco specific nitrosamines (25–32%); phenolics (24–56%), benzene (16%), toluene (25%) and cadmium (34%) with increased yields of increased yields of formaldehyde (49%) and isoprene (17%) (Liu et al., 2011). The research that are conducted using currently available filters to reduce toxicants have focused on modification of filters for improving the filtering capacity and to recycle the cigarette butts and the chemicals trapped in it during filtration (Torkashvand et al., 2022). Development of safest, most reliable and reusable material for filtering the toxic compounds from smoke generated during combustion remains a challenge. This can decrease the health impacts generated by this large industry producing cigarette that are still ineffective to provide full security to smokers against toxic compounds through filters. Studies and reports suggest that polyphenols help to significantly reduces the cytotoxicity, which defines protective role in lung epithelium (Qamar and Sultana, 2011).

According to the study, (Doctype Innovations, 2024) conducted in Nepal, the following are the primary health hazards linked to tobacco use: Non-Communicable Diseases (NCDs): A number of NCDs are significantly influenced by tobacco smoking, including: Diseases of the Heart (CVDs): Smoking raises the risk of stroke and heart disease. Lung diseases include respiratory conditions such as Chronic Obstructive Pulmonary Disease (COPD). Cancer: Smoking is associated with a number of malignancies, most notably lung cancer and cancers connected to tobacco use. Mortality: Approximately 27,100 deaths in Nepal are caused by tobacco use annually, making up 14.9% of all fatalities in the nation. This includes 14.1% of female fatalities and 15.6% of male deaths. Disability Adjusted Life Years (DALYs): Tobacco use contributes to 6% of DALYs for all ages and both sexes, indicating a significant burden of disease and disability caused by tobacco-related health issues. These health risks highlight the urgent need for effective tobacco control measures in Nepal.

E-cigarettes had revolutionized the consumption process of cigarette and have been being consumed for about a decade but it has resulted a manifestation period of two or more decades for the generation of numerous smoking related symptoms and health effects (Balfour et al., 2021, Lucchiari et al., 2020,, Roditis et al., 2015, Bozier et al., 2020, Cohen et al., 2022, Begh and Aveyyard, 2020). The steps taken to examine the long-term effects of vaping on public health have been taken too early. Studies have been done through which comparison of the genotoxic and cytotoxic capacity of mainstream smoke obtained from non-filter 2R4F, CA-filter 2R4F, and carbon-filter 2R4F cigarettes were determined, using the assays: bacterial mutagenic assay and a neutral red cytotoxicity assay (Shin et al., 2009). Epidemiological studies have shown that the log-term consumption of polyphenols offer protective effects against the cancer, cardiovascular diseases, diabetes, osteoporosis and neurogenerative diseases strongly suggest that long-term consumption of diets (fruits, vegetables, tea, and coffee) rich in polyphenols offer protective effects against the development of cancer, cardiovascular diseases, diabetes, osteoporosis, and neurodegenerative diseases. (Rudrapal et al., 2022; Castelli et al., 1981; Teo et al., 2006; Herxheimer et al., 1967; Talcott et al., 1989; Auerbach et al., 1979; Cryer et al., 1976). The observations in research carried out suggest that inhalation of tobacco containing nicotine reduces vascular PGL production causing cardiovascular disease (Nadler et al., 1983).

Nitrosamines as shown in Fig. 1 have been seen to be constituents of food, beverage, air, cigarette smoke, cosmetics and industrial environments and the tobacco specific nitrosamines are also major compounds present in tobacco which are needed to be studied and reviewed well. (IARC 17 1978; Banbury Report 1982; Magee 1996; Lin 1990; Preussmann and Eisenbrand 1984; Preston-Martin and Correa 1989; Tricker 1997; Magee 1989; Tricker et al. 1989; Startin 1996; Loeppky and Michejda 1984; Eisenbrand *et al.* 1996; Scanlan 1999;). Tobacco-specific nitrosamines (TSNAs) actually are group of carcinogens which are generated in tobacco smoke. They get produced from nictotine and related alkaloids during the processing of tobacco and tobacco products. The most common Tobacco specific nitrosamines are: NNN (N'-nitrosonornicotine), NNK ((4-methylnitrosamino)-1-(3-pyridyl)-1-butanone), NAB (N'-nitrosoanabasine) and NAT (N-nitrosoanatabine) (Hoffman et al., 1982). The research conducted over hamster rats showed that NNN and NNK induced tumors in the upper respiratory tract of the rats and that NNK is the most effective and active carcinogen of TSNA that induce adenoma and adenocarcinoma in the human lung. Such harmful effects over the consumer's lung can be predicted in human beings too (Hoffman et al., 1982; Konstantinou et al.,2018; Pool-Zobel et al, 1992; Levy et al., 2004; Arredondo et al., 2006; Ashley et al., 2010). Among nicotine-derived carcinogens, the most important one is nitrosamine is 4-(methylnitrosoamino)-1-(3- pyridyl)-1-butanone (NNK). Secondary reduction of NNK degrades to 4-(methylnitrosoamino)-1-(3-pyridyl)-1-butanol (NNAL) that has an adverse health effect (Matt et al., 2011). So, there is a need of invention to reduce such compounds through functionalization of cigarettes (Hoffman et al., 1995).

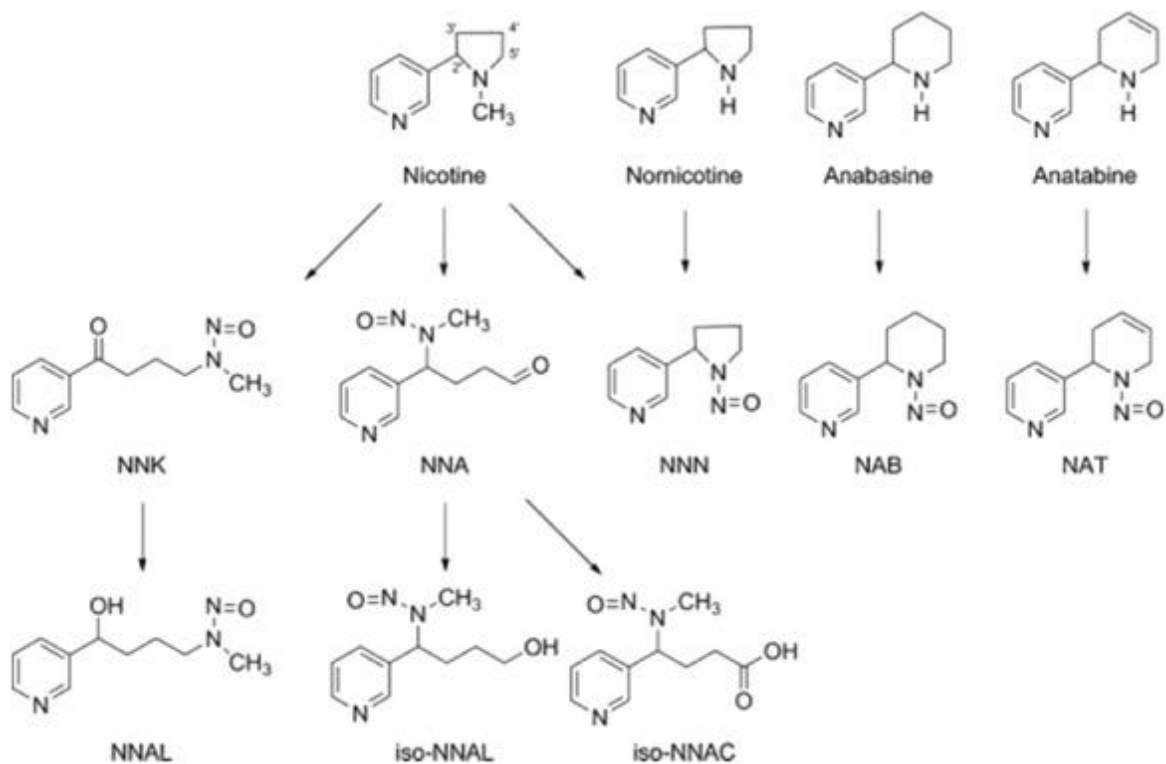

**Fig. 1.** Formation of Tobacco specific N-nitrosamines (Konstantinou et al., 2018).

Some of the most toxic isomers of nitrosamines are iso-NNAC, 4-(methylnitrosamino)-4-(3-pyridyl) butyric acid; iso-NNAL, 4-(methylnitrosamino)-4-(3-pyridyl)-1-butanol; NAB, N′-nitrosoanabasine; NAT, N′-nitrosoanatabine; NNA, 4-(methylnitrosamino)-4-(3-pyridyl)butanal; NNAL, 4-(methylnitrosamino)-1-(3-pyridyl)-1-butanol; NNK, 4-(methylnitrosamino)-1-(3-pyridyl)-1-butanone; NNN, N′-nitrosonornicotine Note: NNA is a very reactive aldehyde and has therefore never been quantified in tobacco or tobacco smoke (Hoffman et al., 1995). Commerical cigarettes as shown in Fig. 2, have high potential of producing such carcinogenic compounds during combustion.

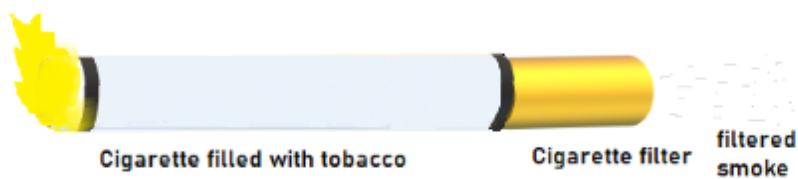

**Fig. 2.** Commercial cigarette.

Murphy et al. (2017) found that the margin of exposure (MOE) values for propionaldehyde, acetone, butyraldehyde, and acetaldehyde were more than 10,000. This implies that certain drugs might be given less priority when it comes to risk management measures. On the other hand, MOE values for formaldehyde and acrolein, which were obtained from EC use, dropped below 10,000, suggesting that there is a need for further focus. The low priority position for risk management actions was further supported by the MOE values exceeding 10,000 for acetaldehyde, propionaldehyde, butyraldehyde, and acetone. On the other hand, based on EC usage, the MOEs for formaldehyde and acrolein were less than 10,000, indicating a greater need for action. It was observed that the MOEs for EC usage were one to two orders of magnitude greater than those linked to smoking exposure for the same toxicants. Moreover, the MOE values obtained from EC emissions for the nitrosamines N-nitrosonornicotine (NNN) and N-nitrosodimethylamine (NDMA) exceeded 10,000. Nevertheless, the MOEs for butyraldehyde, acetone, and NNN were based on little information, mostly from non-inhalation or acute exposure investigations. The study showed that the lower emission levels from EC use led to a less hazardous toxicological profile when compared to traditional cigarette smoking.

Furthermore, smoking was found to be a significant contributing factor in an in-silico research on transitional cell carcinoma (TCC), the most common kind of bladder cancer (Vitti et al., 2020). According to experimental findings from the research conducted on rats (Visioli et al., 2000), the negative effects of smoke-induced oxidative stress were lessened when small amounts of hydroxytyrosol were given through wastewater from olive mills. Additionally, polyphenols derived from olive trees may be used to alter cigarette smoke vapors, lowering nitrosamine levels, according to the research by de Falco et al. (2020). Despite recent increases in awareness of the potential environmental harm that cigarettes and their constituents may cause, prior studies on tobacco toxicity have mostly concentrated on smoking-related illnesses and how to prevent and treat them in people (such as lung cancer) (Santos-Echeandía et al., 2021).

Consumption of e-cigarettes is not a risk-free habit too. In the case study conducted in Kathmandu Metropolitan City (KMC) in (Khanal et al., 2023). The lifetime e-cigarette usage rate was 21.2%, while the current e-cigarette use rate was 5.9%. Having more close friends (AOR = 5.23, CI: 1.26, 16.39), friends who also used e-cigarettes (AOR = 7.23, CI: 0.93, 22.82), being a man (AOR = 2.88, CI: 2.15, 10.35), and being older (COR = 5.07, CI: 0.93, 8.19) were important variables impacting current e-cigarette usage. The number of friends who used e-cigarettes (AOR = 5.90, CI: 2.15, 10.35), male gender (AOR = 3.53, CI: 2.15, 10.35), age (COR = 4.56, CI: 0.98, 6.24), and place of residence (COR: 5.19, CI: 0.83, 8.02) were also associated with lifetime e-cigarette usage. Males were much more likely than females to use it (5.4% vs. 0.5%). While 56.5% of respondents acknowledged that e-cigarettes might increase conventional smoking, 34.8% of respondents saw e-cigarettes as a tool to help people quit smoking. Most undergraduate students (64.7%) thought e-cigarettes were safer than traditional smokes.0.5%. The study (McNeill et al., 2015) successfully demonstrated the important events that connect toxicant exposure to smoking-related disorders, covering a wide range of chemicals, using the Adverse Outcome Pathway (AOP) architecture. This paradigm was used in a number of investigations, including chemical analysis, in silico modeling, in vitro biological evaluations, and human trials, to compare a closed modular e-cigarette (the Vype e-Pen) with traditional cigarettes. The results yielded the most extensive dataset for a single e-cigarette product to date. All things considered, our findings support Public Health England's findings, which indicate that modular e-cigarettes, such as the e-Pen, are less dangerous than traditional cigarettes.

Particulate matter (PM2.5 and PM10), carbon monoxide (CO), sulfur dioxide ($SO_2$), nitrogen dioxide ($NO_2$), ozone ($O_3$), and volatile organic compounds (VOCs) are common airborne hazardous substances that can be harmful when inhaled (Sylvester Chibueze Izah et al., 2024). Respiratory and cardiovascular issues can result from particulate matter (PM), particularly tiny particles (PM2.5), which can enter the circulation and go deep into the lungs. Reduced oxygen supply to bodily tissues can result from the colorless gas carbon monoxide, which is generated by industrial operations and vehicular emissions (Cope, 2020). Sulfur dioxide, which is mostly produced by burning fossil fuels, can aggravate the respiratory system and exacerbate symptoms of asthma. Nitrogen dioxide, which comes from combustion sources as well, can lower immunity to infections and cause inflammation in lung tissue (Krismanuel & Hairunisa, 2024). Smog's main ingredient, ozone, irritates the airways and exacerbates respiratory conditions (Arif & Hassan, 2023). Last but not least, volatile organic compounds, or VOCs, are emitted from a variety of sources, such as paints and gasoline, and can raise the risk of cancer and respiratory problems. Such toxic compounds are blocked by filters used in face maks as shown in Fig. 3.

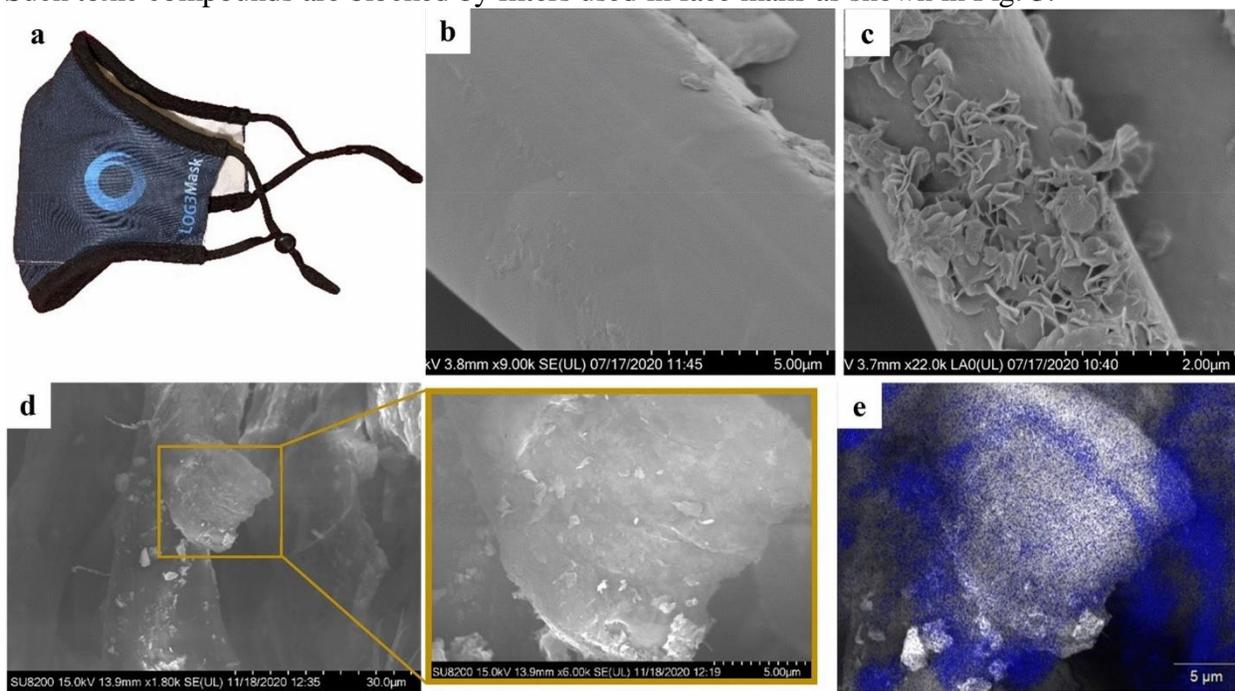

**Fig. 3.** Functioning of face masks (Gonzalez et al., 2021).

The purpose of this review is to compare the processes and discover the most appropriate process of making safer cigarette and mask filter in developing country like Nepal using modern technologies using past studies and inventions. This review critically analyzes features, functions, and properties of various filters and evaluates potential of such filters for application in developing country like Nepal where pollutions and smoking related diseases are common problems.

**2. Materials and methods of making a filter**
A variety of cutting-edge techniques and materials that improve filtering efficiency are used in the manufacturing of filter materials. These techniques provide efficient filters for the filtration of air and water by utilizing a variety of fibers, such as synthetic, biological, and composite materials. Important elements of filter material processing are described in the Table 2.

**Table 2** Progress on filter technologies

| Category | Details | Examples/Materials | Key Advantages | Challenges | References |
|---|---|---|---|---|---|
| Fiber Types | Nanofibers and composite fibers are key components in filter production. | Polyamides, polyesters, PE-PP composite fibers | High permeability, structural integrity | Cost of advanced materials | (Ergashev et al., 2023; Yuru, 2020) |
| Fiber Composition | Nanofibers are 20-200 nm in size, while composite fibers combine PE and PP for enhanced performance. | Nanofibers, PE-PP fibers | Effective pollutant filtration | Scalability issues | (Ergashev et al., 2023; Yuru, 2020) |
| Manufacturing Methods | Layered structures and crosslinked protein fibers are used to create filters with improved capabilities. | Layered nanosilicon composites, protein fibers | Physical and chemical pollutant neutralization | Complexity of processes | (Feng & Gui, 2019; Zhong et al., 2018) |
| Innovations | Methods to prevent pore blocking using solid fillers have been developed to maintain performance. | Solid filling agents | Prevention of pore blockage, consistent efficiency | Cost and material compatibility | (Lin & Tao, 2017) |
| Applications | Filters are used in air and water purification, with enhanced functionality | Air filters, water purifiers | Improved filtration efficiency | Limited large-scale adoption | (Ergashev et al., 2023; Yuru, 2020; Feng & Gui, 2019; Zhong et al., 2018; |

| | | | | | Lin & Tao, 2017) |
|---|---|---|---|---|---|
| Future Focus | Optimization of processes to address cost and scalability for broader applications. | Advanced materials and techniques | Broader applications in various industries | Economic and technical limitations | (Ergashev et al., 2023; Yuru, 2020; Feng & Gui, 2019; Zhong et al., 2018; Lin & Tao, 2017) |

*2.1. Inventions of filters based on materials and methods*

There were different materials and methods used in the past in different studies to create filters as discussed in Table 3.

**Table 3** Summary of materials and methods for making filters.

| Section | Objective/Function | Materials/Methods | Results/Findings | References |
|---|---|---|---|---|
| Cellulose Acetate for Filters | Examine cellulose acetate filters for cigarette applications. | Stainless steel tube, Teflon capsule, spectrometer, triacetate/secondary acetate fibers, calcium chloride, acetic acid. Airflow was drawn at 17.5 ml/sec; sorption power assessed, and analysis via spectrophotometry or titration. | Efficient sorption power and air resistance measured; fibers spun using acetic acid in an aqueous bath. | (Markosyan et al., 1971) |
| Functional Nanocoating | Investigate plant-derived phenolic compounds for nanocoating filters. | Phenolic compounds (e.g., pyrogallol, gallic acid, catechin) with antioxidants (sodium ascorbate, glutathione, uric acid). Nanocoating slowed oxidation kinetics of phenolic compounds. | Improved stability of phenolic compounds through antioxidant-assisted nanocoating. | (Sadabad et al., 2019) |

| Topic | Objective | Methodology | Findings | Reference |
|---|---|---|---|---|
| Pd and Ni Loaded Carbon Nanotubes | Remove tobacco-specific nitrosamines (e.g., NNK) using Pd/Ni-loaded SWCNTs. | Gaussian 09 software for modeling SWCNT (70 carbons, 20 hydrogens). Ni atom substituted in SWCNT structure. Adsorption interactions evaluated with density functional theory, binding energy calculations, and natural bond orbital (NBO) analysis. | High adsorption efficiency for NNK on Pd/Ni-loaded SWCNT. Optimized geometries and electronic interactions demonstrated strong binding capabilities. | (Yoosefian, 2018) |
| Toxic Carbonyl Species Removal | Reduce toxic carbonyl species in e-cigarettes using reactive carbonyl species (RCS) trapping. | Gallic acid, hydroxytyrosol, and epigallocatechin prepared at varying concentrations (0.6–5 mM). Aerosols were collected using Subox Mini C device. Analysis via HPLC-UV and LC-MS to evaluate dicarbonyl adduct formation. | Effective trapping of toxic carbonyl species; reduced cytotoxicity in e-liquid aerosols. | (de Falco et al., 2020) |
| Polyphenol Functionalization | Develop filters functionalized with polyphenols for enhanced medical mask filtration. | Cellulose wipes with catechin (>98%), laccase enzyme, sodium tartrate buffer, and Tween 80. Grafting performed at 50 °C with 4-hour incubation, followed by water washing and air drying. | Successful grafting of polyphenols onto cellulose with improved filtration efficiency. | (Catel-Ferreira et al., 2015) |

.

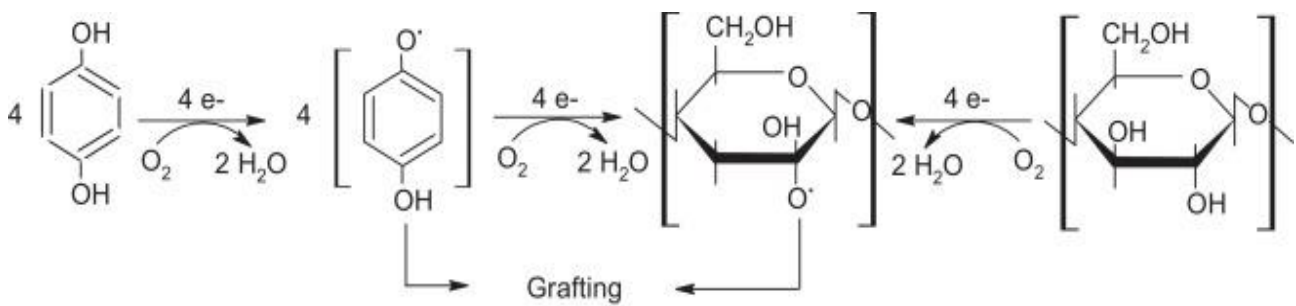

**Fig. 4.** Mechanism of action of laccase enzyme on polyphenol and cellulose during grafting. (Catel-Ferreira et al., 2015).

*2.2. Inventions of modern filters*
Moreover, the review of latest findings and inventions on filter manufacturing process are summarized in Table 4.

**Table 4** Latest inventions and discoveries in filter making technologies.

| References | Discoveries |
|---|---|
| (Lin, 2020) | Through the processes of fiber web formation, coating, drying, and papermaking, the filter material is created from hot melt composite fibers (PE skin, PP core) and self-twisted short fibers (40–50 mm). |
| *(Verstraete & Proost, 2021)* | The filter element is made using a pleated or fluted filter media, and the seal arrangement is made using multi-material injection molding and thermal welding. |
| *(Gao et al., 2019)* | The second physical filtering layer of the filter material is comprised of activated carbon, which is joined to a chemical filtering functional layer by a connecting layer that is either ultrasonically formulated or sprayed with adhesive. |
| *(Damian, 2020)* | The process creates optical filter materials appropriate for a range of uses, such as augmented reality and laser protection, by depositing chiral nematic liquid crystal onto a substrate and aligning it under pressure. |
| (Hanley et al., 2020) | The techniques for creating and utilizing the fibrous material, bonding material, and flame-resistant treatment composition that make up the filter media are also described. |

| (Zhong et al., 2018) | Air filters are made by assembling crosslinked protein-containing nanofibers into mats, which allow contaminants to be chemically filtered while allowing air to pass through. |
|---|---|
| (Zhong et al., 2018) | Protein-containing nanofiber mats are used to create air filters by forming nanowires into a mat structure with holes for airflow and filtration. |
| (Cordatos et al., 2018) | In order to minimize carbonaceous deposits, the filter screen consists of a plate body with a variety of holes and a layer of polytetrafluoroethylene distributed conformally on the upstream surface. |
| (Matteucci & Napolitano, 2015) | A convergence device for the filter material, a thread positioning device for continuous thread placement, and a liquid introduction device for adding liquid to the filter material are all part of the process. |
| (Medvedev & Karapetjan, 2015) | Electrospinning is used to create fibers from solutions of two polymers, which, when they come into contact, acquire a triboelectric charge that strengthens the material's electric field and filtering capabilities. |
| (Elrod, 2004) | The process entails depositing different drop sizes onto a substrate framework to construct a cell utilizing an inkjet print head and a liquid filter material ejection device. |
| (Tanaka et al., 2002) | After adding a liquid lubricant to finely ground PTFE powder, the tape is formed, stretched to create a porous layer, and heated to create an air-permeable support member. |
| (Tueshaus & McGrenera, 2006) | The filter is constructed from spirally wrapped strips of filter material, including wire mesh, that are joined by resistance welding to allow for a range of lengths. |
| (Tate et al., 2020) | Using a filter media sheet, non-roll stamping dies are used to press embossments onto the sheet, which is then pleated and collected to create a pleated filter media pack. |
| (McGaw, 2000) | As part of the production process, many layers of filter material are assembled, and the edges are concurrently chopped and sealed using ultrasonic welding. |

## 3. Progress in filter technologies

In the early 1930s, cigarettes with filters were created to enhance the smoking experience and lessen the harm that chemicals caused to the human body and progressively embraced by smokers (Cao et al., 2023). There has been gradual progress in filtration technologies of filtering membranes used in masks and cigarettes. Some recent progresses and insights on development have been discussed in different subsections.

*3.1. Recent development of filters in cigarettes*

The goals of recent advancements in cigarette filters have been to improve user experience, sustainability, and filtering efficiency. The utilization of biological materials, sophisticated filtration systems, and creative designs that enhance smoke quality and lower hazardous emissions are examples of innovations. These developments are a result of increased knowledge of the negative effects smoking has on one's health and the environment.

If biological materials are considered, by using vegetable flour paste to create compartments for both smoke flow and filtration, a novel filter design has the potential to improve the environmental impact of cigarette manufacture (Alessandro, 2021). Advanced Filtration Systems to improve the characterisation of secondhand smoke, researchers have created a filter-based system that simulates the deposition of respiratory particles in humans. High repeatability and particle-filtering effectiveness were displayed by this technology, which is essential for comprehending the health hazards linked to smoking (Hao et al., 2022).

In order to increase hardness and decrease tar visibility, recent designs use filters with different smoke channel diameters, improving user experience without compromising structural integrity (Yu & Kenichi, 2014). Furthermore, a filtering unit enables the dynamic modification of filtering materials, which may enhance taste and filtration efficiency while in use (Shaoqiang, 2015). On the other hand, even though these developments are meant to lessen health concerns, smoking still has intrinsic risks, and it is still up for discussion whether these filters can actually considerably reduce harm. Table 5 below presents the literature review on recent progress in development of filters:

**Table 5** Literature review on recent progress in development of cigarette filters.

| References | Literature review |
|---|---|
| (Conradi et al., 2019) | The research focuses on making adsorbents from cigarettes to remove lead from water, however it ignores current advancements in cigarette filters. |
| (Seng, 2013) | Recent developments include filter media with continuous nanofiber or submicron fiber, featuring low air resistance and specific inter-fiber spacing, enhancing performance in cigarettes and related products. |
| (Shin et al., 2014) | One recent innovation is the use of bamboo fibers in cigarette filters, which increase suction resistance, manufacturing workability, and minimize paper odor by using a dry crimping technique. |

| (Antonella et al, 2015) | With the use of a radiofrequency generator, the device outlined speeds up the aging process of cigarette filters while improving humidity removal through a variable residence duration in the drying chamber. |
|---|---|
| (Szu-Sung et al., 2008) | In order to decrease particle breakthrough and improve the effectiveness of smoke treatment, recent innovations include activated carbon adsorbents and filters containing electrostatically charged fibers. |
| (Kim et al., 2009) | Cigarette filters made of paper coated with flavored natural plant extracts are one recent innovation that improves the elimination of toxic substances from cigarette smoke. |
| (Donato et al., 2023) | This study highlighted the possibility of functionalizing cigarette filters with olive polyphenols, namely hydroxytyrosol, to reduce the production of toxic substances, especially nitrosamines, when smoking. A two-layer capture factor (f = 2.9×103) demonstrated the exceptional filtering performance of functionalized filters. |

*3.2. Filters used in medical masks*

To improve its ability to stop the spread of airborne infections and particulate matter, medical masks use a variety of filters. These filters' overall performance and filtering efficiency are greatly influenced by their construction and composition. The main features of the filters used in medical masks are listed in Table 6.

**Table 6** Review of features of the filters in medical masks.

| Category | Description | Source |
|---|---|---|
| **Filtration Efficiency** | Medical masks (ASTM-certified level 1 and level 3) demonstrate filtration efficiencies of 52% to 77% against small aerosols (0.02 - 1 μm). Advanced masks like N95 and CaN99 achieve filtration efficiencies of 97-99%, suitable for high-risk environments. | (Tomkins et al., 2024) |
| **Innovative Filter Designs** | A high-filtration composite medical mask incorporates an activated carbon layer between filter cotton yarns for enhanced filtering. | (Songbo, 2020) |
|  | The medical efficient aerosol filtering mask features multiple layers (primary filtering and moisture absorption) to improve aerosol filtration and biological matter inactivation. | (Minqiang, 2018) |
| **Alternative Materials** | Research has explored bacterial cellulose combined with natural extracts to create filter membranes with 99.83% efficiency in removing particulate matter and antimicrobial properties. | (Jonsirivilai et al., 2022) |

Detailed literature review on medical filters are mentioned in Table 7.

Table 7 Literature review on medical mask filters.

| References | Literature review |
|---|---|
| (Liang et al, 2022) | An excellent substitute for medical mask filters, the SAg air filter is composed of sodium alginate and nanosilver and provides outstanding air filtration, antibacterial qualities, and biodegradability. |
| (Wu et al., 2023) | In order to achieve high filtration effectiveness and dust holding capacity appropriate for medical mask applications, the study introduces a multilayer composite membrane that combines PTFE nanofiber layers with PP melt-blown layers. |
| (Goodge et al., 2024) | Medical mask filters can benefit from hybrid PVA nanofibers that combine PB and PMA to improve filtration efficiency and preserve electrostatic capture. |
| (Im & Hong, 2014) | Hydro charging improved filtering without compromising breathing resistance, while a melt-blown nonwoven filter for medical masks demonstrated greater filtration efficiency with increased basis weight. |
| (Martí et al., 2021) | More than 99% of SARS-CoV-2 and multidrug-resistant bacteria are successfully rendered inactive by a new non-woven face mask filter coated with benzalkonium chloride, improving protection in hospital environments. |
| (Le et al., 2022) | The PLLA nanofiber filter offers over 99% efficiency for PM 2.5 and is reusable, humidity-resistant, and biodegradable, making it suitable for medical masks. |
| (Choi et al., 2021) | With a 98.3% removal rate of 2.5 μm particulate matter, the biodegradable, moisture-resistant, and highly breathable filter created in this study is just as effective as N95 filters. |
| (AmirHooman SadrHaghighi et al., 2023) | The study created a melt-blown filter for facemasks coated with copper nanoparticles, which showed antiviral and antibacterial qualities, improving defense against airborne illnesses such as SARS-CoV-2. |
| (Abbas et al., 2021) | Medical mask filters can benefit from the excellent filtering efficiency and comfort of |
| | novel electrospun composite layers that include TiO$_2$ nanotubes and chitosan/poly (vinyl alcohol). |
| (Cheng et al., 2022) | In order to improve the filtration performance of medical masks against hazardous aerosols, nanoporous atomically thin graphene screens may efficiently block sub-20 nm aerosolized nanoparticles. |
| (Jonsirivilai et al., 2022) | The study offers a filter membrane made of biodegradable bacterial cellulose and fingerroot extract that removes 99.83% of particle matter, making it appropriate for use in medical masks. |
| (Choi et al., 2021) | A high-performance, biodegradable fibrous mask filter was developed, combining biodegradable microfibers and nanofibers with a chitosan nanowhisker coating. This innovative filter effectively captured 98.3% of fine particles and maintained a comfortable breathing pressure, outperforming commercial N95 filters in moisture resistance and usability. Additionally, it fully biodegraded in compost soil within four weeks, making it an environmentally friendly alternative. |
| (Zortea et al., 2024) | In order to increase filtering efficiency, this study investigates the use of cellulose nanofibrils (CNFs) derived from sugarcane bagasse (SCB) in nonwoven face masks. A simple alkaline pretreatment (CNF-PA) and an extra enzymatic pretreatment (CNF-AE) were the two CNF kinds that were created. In comparison to CNF-PA, CNF-AE films showed better qualities in terms of contact angle, density, porosity, transparency, and vapor permeability. The bacterial filtration efficiency of masks coated with CNFs employing a spray approach were 99.68% (CNF-PA) and 99.80% (CNF-AE), above the 95% minimum requirement. Breathability was also good, demonstrating that SCB CNFs can improve mask protection and filtration. |

According to Choi et al. (2021), the process for creating the effective and biodegradable face mask filter included a number of crucial processes. In order to increase filtering efficiency, the first step in the material selection process was to employ poly (butylene succinate) (PBS) as the main component, supplemented by chitosan nanowhiskers (CsW). In order to produce a fiber mat with certain morphologies and pore sizes, solutions of different concentrations were produced and electrospun as part of the production process. To improve their electrostatic filtering properties and maximize contact with particulate matter (PM), the electrospun PBS fibers were coated with CsW after production. This technique imparted permanent ionic charges. Scanning electron microscopy (SEM) and other analytical methods were used to analyze the physical characteristics of the resultant mats, such as fiber diameter, pore size, and base weight. Under controlled airflow settings, filtration testing was carried out to evaluate the effectiveness of PM removal for different particle sizes (PM1.0, PM2.5, and PM10). Performance was assessed using the pressure drop across the filter and the percentage of PM removed. Furthermore, composting experiments and enzymatic degradation tests were used to determine weight loss and deterioration over time as part of biodegradability assessments. In order to make that the new filter maintained high filtering requirements while offering a comfortable breathing environment, its performance was finally compared to that of traditional filters, including N95 masks, in terms of pressure drop and filtration efficiency. This thorough process produced a high-performing, biodegradable mask filter that was both ecologically benign and efficient at capturing particle matter.

As discussed in Table 7, the research (Choi et al., 2021) has presented a more favourable invention applicable in developing countries even in Nepal. As shown in Fig. 5, burning a cigarette produced a significant amount of PM; the filter, placed between the PM (particulate matter) source and another empty box, was visually verified to totally prevent the created smoke from entering the empty box via the Tyndall effect. Through electrostatic interactions, the nanofiber side adsorbs relatively tiny PM after the microfiber side that faces the smokey chamber sieves relatively big PM.

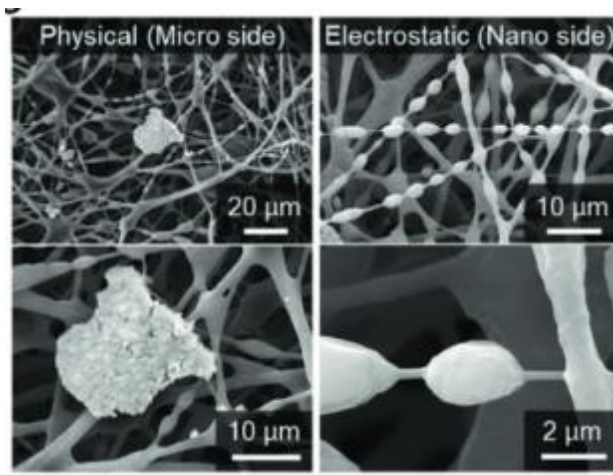

**Fig. 5.** Images displaying the physical and electrostatic PM capturing potentials of the integrated filter (Choi et al., 2021).

As discussed in (Choi et al., 2021), one of the study's distinctive features is that the very efficient filter is made entirely of biodegradable materials. Given the enormous number of masks already in use, there will surely be a serious waste issue in the near future. The lipase enzyme from Thermomyces lanuginosus completely degraded ChN2.5 in 7 hours at 50 °C, as shown in Figure 5a and Movie S3, Supporting Information. Furthermore, in composting soil, it decomposed entirely in four weeks at room temperature. Therefore, biodegradable materials, especially CsW and PBS, provide obvious substitutes for mask waste management. The performance promises of the biodegradable filter are supported by Fig/ 5. It highlights that because of its minimal pressure drop, the new filter not only effectively removes dangerous particulate matter but also offers a comfortable experience. For the comfort of the user, the biodegradable filter is made to maintain a minimal pressure drop. The figure could show that although the biodegradable filter has a high filtering efficiency, it does it with a much smaller pressure drop than traditional filters, which makes it more comfortable to use for extended periods of time. The biodegradable filter's excellent efficacy and comfort make it a competitive alternative to standard masks, especially in urban or smoke-polluted situations like those smokers' experience. The study's main conclusions are summarized in Fig/ 6, which also highlights the benefits of the novel filter and supports its potential for real-world use in PPE.

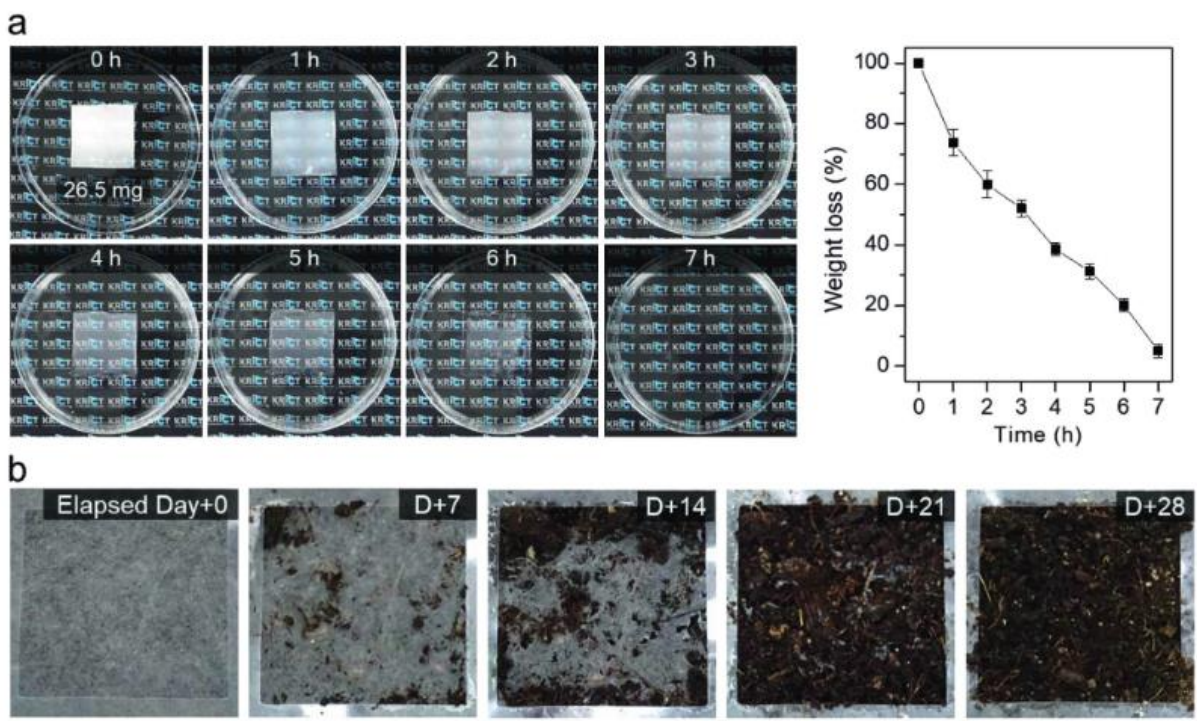

**Fig. 5 (a).** Images demonstrate the CsW-coated PBS filter's time-dependent enzymatic breakdown and the associated weight loss, **5( b).** Pictures illustrating the CsW-coated PBS filter's deterioration over time in the composting soil (Choi et al., 2021).

*3.3. Effects of development of modern filters*
Modern filters in cigarettes and masks have greatly increased their ability to shield users from dangerous gasses and particles. Safety, user comfort, and filtering effectiveness have all increased as a result of advancements in filter materials and designs. This development may be divided into three categories: improvements in cigarette filters, mask filters, and gas masks and respirators. The advancement on the filtering technologies are effect of such recent developments which are described in brief in Table 8.

**Table 8** Effects of progress of filter technologies.

| Category | Advancement | References |
|---|---|---|
| Gas Masks and Respirators | | |
| Filter-Sorbing Materials (FSM) | New FSMs combine polymer or glass fibers with activated carbon (AC) to enhance protection against aerosols and gases, including ammonia and inorganic vapors. | (Talipova et al., 2023) |
| Low Resistance | Small-sized filters with low airflow resistance improve breathability while maintaining high filtration efficiency. | (Talipova et al., 2023) |
| Mask Filters | | |
| Bactericidal Effects | Copper wire wrapped around resin fibers discharges static electricity, inactivating bacteria and viruses to improve protective capabilities. | (Kodera et al., 2010) |
| Broad-Spectrum Filtration | Gel breaker agents in filters remove particulate matter from PM 10 to PM 0.1 while preserving mask permeability. | (Shuhua et al., 2015) |
| Cigarette Filters | | |
| Moisture Control | New drying processes optimize moisture levels in cigarette filters, enhancing efficiency and safety. | (Antonella & Monzoni, 2018) |
| Reduced Harmful Emissions | Advanced cigarette filters decrease carbon monoxide levels while continuing tar reduction for a healthier smoking experience. | (Shi et al., 2019) |

*3.4. Insights on development of safe filters*

For the sake of public health, proper filters for masks and cigarettes are essential, especially when it comes to reducing the hazards of airborne infections and dangerous chemicals. Innovative methods to improve filter safety and efficacy have been highlighted by recent researches as mentioned in Table 9, which can help guide policy choices in these areas.

**Table 9** Insights of development of safe filters

| Category | Advancement | References |
|---|---|---|
| Enhanced Mask Filters | | |
| Modification Techniques | Surgical mask filters can be enhanced by coating with sodium dihydrogen phosphate, capturing 40-60% of aerosolized viruses and inactivating them, which significantly reduces viral infectivity. | (Lee et al., 2021) |
| UV-C Integration | Integrating UV-LEDs in masks allows for the disinfection of inhaled and exhaled air, reducing microbial contamination and extending the mask's lifespan. | (Christoph & Chih, 2020; Haitao, 2017) |
| Cigarette Filter Safety | | |
| Material Concerns | Cigarette filters, largely made from cellulose acetate, can release respirable fibers that resist biodegradation and may carry carcinogenic tar, posing potential health risks. | (Hastrup et al., 2001) |
| Consumer Awareness | Increased consumer education on the inhalation risks of filter fibers could aid in implementing regulatory measures for public safety. | (Hastrup et al., 2001) |

Moreover, a literature review on safety considerations in filtration systems developed in cigarettes and masks have been described in Table 10.

Table 10 Literature review on safety considerations.

| Literature review | References |
|---|---|
| In order to improve safety against infections, the study highlights the necessity of creating antimicrobial filters for masks. It also suggests that comparable advances may be investigated for cigarette filters in order to lessen hazardous exposure. | (Martí et al., 2021) |
| The study focuses on a cigarette filter that collects smoke, and it makes the argument that comparable processes might improve mask filter safety by keeping dangerous particles from escaping. | (Brian, 1998) |
| In order to improve safety in the creation of filters for masks and cigarettes, the study focuses on a disposable filtering mask that blocks hazardous particles. | (Moscatelli, 2010) |
| Similar advancements in cigarette filters may be influenced by the study's emphasis on the necessity of stringent safety assessments and legal frameworks to guarantee the effectiveness and security of antimicrobial filters in masks. | (Martí et al., 2020) |
| Policies for safer mask and cigarette filter designs may be influenced by the study's recommendations for creating filters that are modeled after animal noses and emphasize low resistance and high particle capturing effectiveness. | (Yuk et al., 2022) |
| The study focuses on creating biodegradable nanofiber membranes for mask filters, and it makes the case that, using sophisticated electrospinning processes, comparable materials might improve cigarette filters' performance and safety. | (Wang et al., 2022) |

| Significant differences in filter efficiency are highlighted by the study, which suggests that regulatory criteria be revised to guarantee reliable performance and safety in protective gear. | (Carlos et al., 2012) |
|---|---|
| The study highlights the necessity of using environmentally friendly, biodegradable materials, such as bacterial cellulose, in the construction of filters because of their effectiveness in removing particulate matter and their antibacterial qualities, which are crucial for the protection of public health. | (Jonsirivilai et al., 2022) |
| Policy ideas for the creation of safe filters in masks and cigarettes are not covered in the report. | (Shah et al., 1983) |
| The research lacks insights on that particular problem since it concentrates on standardizing cigarettes to lessen their attractiveness and addictiveness rather than safe filters for masks or cigarettes. | (Yvette, 2022) |
| The past abuse of filters by the tobacco industry emphasizes the necessity of strict, independent research and regulation to guarantee that any advancements in filters actually lower health risks without deceiving customers. | (Ransom et al., 2021) |

## 4. Environmental effects of cigarette and masks

As the filtering membranes in cigarettes and masks are essential to be used, but they can have harmful effects on environment which need to be considered before applying them in such devices. Cigarette and mask filters have a big impact on the environment, especially when it comes to public health and air quality. While mask filters may lose their effectiveness when exposed to cigarette smoke, cigarette filters produce harmful substances that add to pollution. This interaction calls into question the effectiveness of preventative measures as well as direct exposure to dangerous compounds. The significant pollution from cigarette and mask filters affects aquatic ecosystems and air quality. Cigarette filters, which are mostly made of cellulose acetate, are the most littered item in the world, with an estimated 7 billion thrown away annually (Kadir & Sarani, 2015). These filters discharge toxic substances into the environment, putting human health and marine life at risk (Nitschke et al., 2023). However, mask filters, particularly those exposed to cigarette smoke, show a marked decline in filtration capacity, potentially leading to increased exposure to harmful pollutants (Heo et al., 2022). Literature review on pollution caused by face masks and cigarettes are discussed in Table 11.

**Table 11** Literature review on pollution caused by cigarettes and face masks

| References | Literature review |
|---|---|
| (Guo et al., 2023) | The study does not particularly address pollution from masks or cigarettes; rather, it focuses on evaluating individual exposure to harmful chemicals in ambient tobacco smoke. |
| (Chan et al., 2021) | Polyurethane masks are used in the study to measure exposure to tobacco smoke in the environment; pollution from masks or cigarettes is not the main emphasis. |
| (Beutel et al., 2021) | Masks may also produce comparable waste, which can affect ecosystems through leachates and pollutants in soil and water. Cigarette trash is one source of plastic and microplastic pollution. |
| (Lal & Swaroop, 2017) | With its many harmful substances, cigarette smoke greatly adds to indoor pollution, and masks can pollute the environment by being improperly disposed of, which can result in microplastic contamination. |
| (Wang et al., 2023) | While cigarette pollution is not discussed, the article concentrates on face mask pollution and emphasizes the environmental risks associated with it. Although they are handled differently, both have a major impact on environmental deterioration. |
| (Jiang et al., 2024) | The study emphasizes how cigarette pollution impacts soil health and ecological balance similarly to how mask wastes bind contaminants, preventing plant development and upsetting soil microbial populations. |
| (Fernández-Arribas et al., 2021) | According to the report, cigarette waste presents serious environmental risks, and COVID-19 face masks may introduce organophosphate esters, which might contribute to pollution. |
| (Amuah et al., 2022) | The study emphasizes how cigarette waste offers serious environmental risks and raises pollution levels overall, while worn face masks add to plastic pollution and release toxic compounds. |
| (De-la-Torre et al., 2023) | The study ignores cigarette pollution in favor of concentrating on the pollution generated by single-use face masks, emphasizing their role in microplastics and environmental contamination. |

According to P. Kattel et al. (2023), the COVID-19 pandemic significantly increased the production and use of disposable face masks, exacerbating the global waste problem. Initially used by healthcare professionals, the widespread adoption of these masks by the public raised concerns about proper disposal, particularly in underdeveloped nations like Nepal, where misconceptions about mask composition were common. Fourier Transform Infrared Spectroscopy (FTIR) analysis conducted by the authors on commonly used masks in Nepal's Kathmandu Valley revealed that the primary material was polypropylene microplastic polymers, rather than biodegradable fibers. The Fig. 6 shows an example of the pollution caused by face masks in kathmandu during the period.

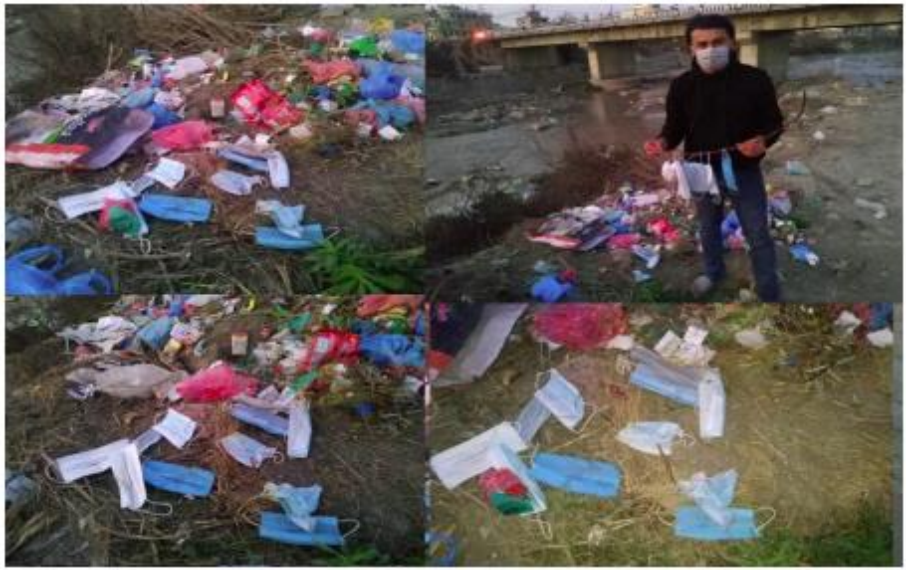

**Fig. 6.** Pollution caused by face masks in Kathmandu (Kattel et al., 2023)

According to retail survey conducted in (John & Ross, 2017), 4,307 empty cigarette packages were collected from 1,204 single stick stores that were tested (see Table 12). The merchants fell into five categories: general stores (55 percent), which included mini-marts and grocery stores; hotels and restaurants (28 percent); paan shops (seven percent); tea shops (eight percent); and other sorts (two percent), which included bakeries, dairies, and cigarette-selling electronic stores. Although all participating merchants agreed to participate in the study, 48 retailers were excluded from the analysis of empty packs since they did not have any packets collected on the day of the survey because of poor sales.

**Table 12** Number of cigarette retails in Nepal in 2024.

| Province | Number of Retailers |
|---|---|
| Province 1 | 171 |
| Madhesh | 204 |
| Bagmati | 209 |
| Gandaki | 95 |

| Lumbini | 170 |
|---|---|
| Karnali | 59 |
| Sudurpaschim | 96 |
| Kathmandu | 200 |
| Total | 1204 |

The study (John & Ross, 2017) dispels the myths that cheap, low-quality cigarettes are illegally imported because domestic cigarettes are heavily taxed, that illicit cigarette sales are large because of the porous border with India, and that the price is the only factor influencing illicit commerce.

A popular source of forecasting the cigarette smokers by country (*Tobacco Products - Nepal | Statista Market Forecast*, 2024) forecasted the cigarette smoking related information for Nepal as presented in Fig. 7 and Fig. 8 which predicted rise in tobacco use in Nepal until 2029. The main cause of this trend is cigarettes, whose consumption is expected to increase consistently throughout the course of the projection period. Although smoking tobacco and cigars also exhibit a rising tendency, their share of the total rise is not as significant. Because tobacco is linked to a number of illnesses, including as cancer, heart disease, and lung conditions, this rising usage raises serious public health issues. It also has economic ramifications, which might affect productivity and raise healthcare expenses. Strong tobacco control measures, such as youth prevention programs and cessation programs for existing smokers, are necessary to address this issue. Obviously, this quantity of cigarettes' filters can pollute the environment if disposed haphazardly which is a common issue in developing country like Nepal.

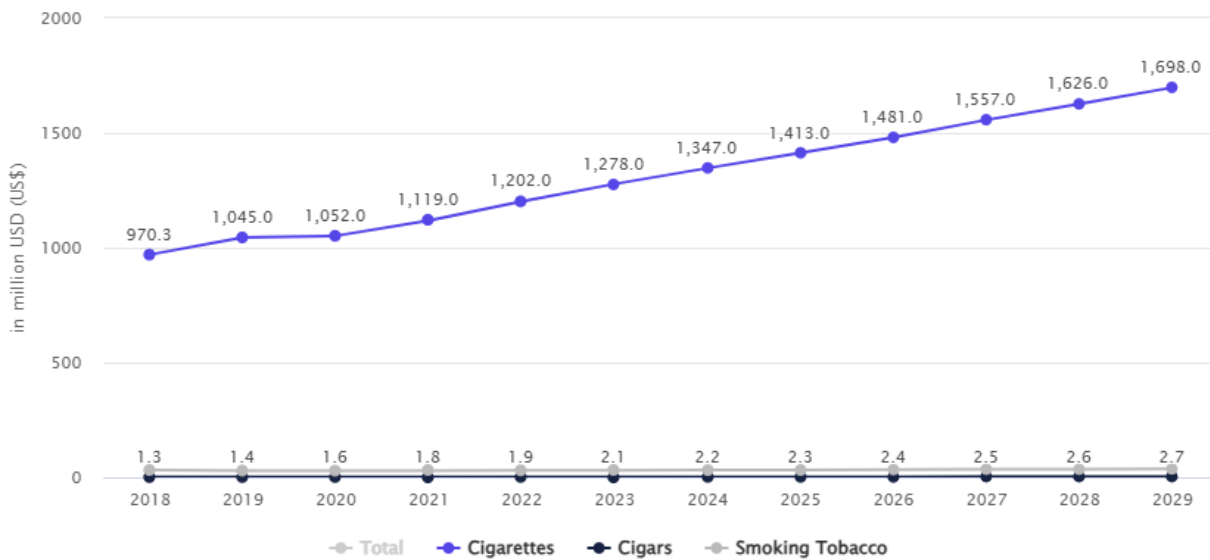

Fig. 7. Curve to represent Cigarettes, cigars, and tobacco consumption in Nepal till 2029.

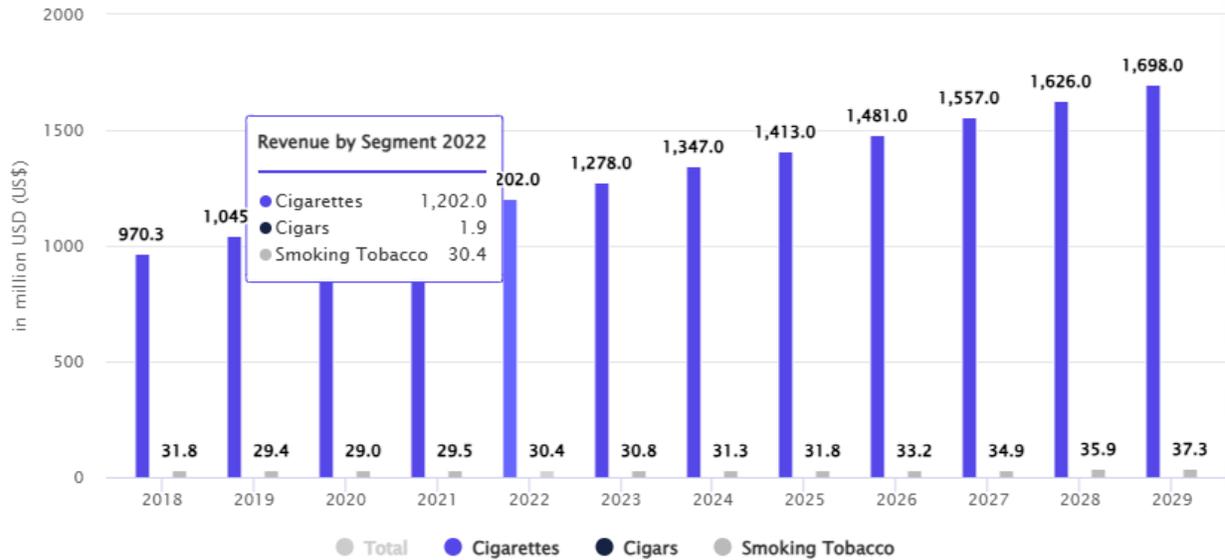

Fig. 8. Bar graph to represent Cigarettes, cigars, and tobacco consumption in Nepal till 2029.

A multifaceted strategy is needed to address the environmental and health effects of toxic cigarettes and pollution from single-use face masks and cigarette trash. To lessen soil and water contamination from cigarette waste, stronger laws governing the biodegradability and disposal of cigarette filters are necessary, as is public awareness of proper disposal practices. The creation of environmentally suitable filter substitutes may also aid in reducing pollution from hazardous chemicals and plastic. According to research by Beutel et al. (2021) and De-la-Torre et al. (2023), reducing mask pollution entails creating reusable masks or single-use masks made of biodegradable materials to reduce microplastic waste and leachate production. The accumulation of plastic debris in natural habitats can be further decreased by providing education and incentives for the appropriate disposal of masks. Global cooperation on regulations to efficiently manage trash disposal and encourage the use of sustainable materials in product manufacture will help both challenges.

## 5. Machine Learning approach:

The accuracy and effectiveness of air quality forecasting can be greatly increased by using machine learning (ML) to predict the Air Quality Index (AQI) for Nepal's air quality data. Large datasets containing a variety of environmental factors, including temperature, humidity, wind speed, and particulate matter (PM2.5, PM10), which have a direct impact on AQI levels, can be processed by ML models. Particularly in areas like Nepal with high pollution levels, these models can offer scalable, real-time predictions that assist individuals and local authorities in making well-informed decisions to safeguard health. Furthermore, by detecting intricate correlations between factors that conventional statistical approaches might overlook, machine learning (ML) improves the accuracy of AQI prediction.

When air quality levels become dangerous, automated machine learning predictions can minimize human error, save time, and provide tailored alerts to sensitive populations, such the elderly and children. Better public health interventions and more proactive steps to lessen the negative impacts of air pollution may result from this.

Machine learning makes a substantial contribution to the study of environmentally friendly cigarette and mask filters as well as explainable AI-based air pollution analysis by providing predictive models that can calculate AQI values, which direct the creation and application of these environmentally friendly filters. It is easier to assess how well these filters reduce dangerous airborne particles like PM2.5 and PM10 when AQI values may be predicted.

The ML models can offer clear insights into how the filters function in actual situations by incorporating explainable AI, which will help Nepal find more sustainable ways to reduce air pollution. As a result, machine learning not only helps to better understand the dynamics of air quality but also offers useful information for enhancing environmental sustainability and public health. An improved approach that increases precision, resilience, and predictive power is the use of machine learning (ML) in conjunction with DNI and DHI for GHI prediction. While classical calculations perform well under ideal circumstances, machine learning (ML) is a potent supplementary tool in solar energy forecasting because it can manage missing data, correct sensor biases, anticipate future values, and account for complex atmospheric interactions. Conventional calculating techniques mostly rely on DNI, which can vary greatly in overcast or hazy conditions, and assume perfect atmospheric conditions. In situations when DNI-based computations alone would be imprecise, ML models can improve forecasts by learning intricate nonlinear correlations between GHI and other atmospheric factors.

*4.1. Dataset collection and feature engineering*

The dataset obtained from an extracted from Environmental Protection Agency (EPA's) repository contained the columns related to time and sensor cleaning classes were eradicated due to their lack of significance in model development. No exclusive feature engineering was needed as data was good clean enough for models' development to predict GHI. The dataset consisted up of (32151 rows, 4 columns).

*4.2. ML Models*

*4.2.1. CatBoost:* It is a gradient boosting technique specifically designed for categorical data. With features like oblivious trees and symmetrical decision splits, it effectively manages categorical features without requiring explicit preprocessing, lowering overfitting and enhancing efficiency.

*4.2.2. Extreme Gradient Boosting, or XGBoost:* It is a potent machine learning technique that uses decision trees as its foundation. It excels in speed and scalability for structured data jobs by combining regularization techniques with gradient boosting to reduce overfitting and maximize performance.

*4.2.3. Extra Trees Regressor:* It is a randomized decision tree-based ensemble learning technique. Its random node splitting, in contrast to conventional tree-based models, offers substantial variance reduction, quick calculations, and predictive accuracy.

*4.2.4. Random Forest Regressor:*

For regression tasks, the Random Forest Regressor is an ensemble machine learning method. During training, it constructs several decision trees and aggregates their results to generate predictions. It lowers the possibility of overfitting and increases the accuracy and resilience of the model by averaging the predictions of individual trees. Large datasets with noise or missing values, as well as non-linear correlations, are especially well-suited for Random Forest. Additionally, it offers feature relevance rankings, which aid in locating the data's most important predictors. All things considered, it is a robust and flexible technique that is frequently applied to predictive modeling jobs.

*4.2.5. Nested Cross validation:* A reliable method for assessing machine learning models and fine-tuning hyperparameters is nested cross-validation. Two loops are involved: an inner loop that cross-validates the training set to adjust hyperparameters and an outer loop that divides the data into training and test sets to evaluate model performance.

*4.3. Evaluation Metrics of the ML model*

*4.3.1. Root Mean Square Error (RMSE):* It the performance indicator of regression model that can be calculated using equation (2) which measures the mean difference between the predicted value by the model with the actual one. It provides the prediction quality of the model.

$$\text{RMSE} = \sqrt{\frac{1}{n}\sum_{a=1}^{n}(x_a - y_a)} \tag{1}$$

*4.3.2. R-squared score:* A statistical model's R-squared ($R^2$) score that can be calculated using equation (3) indicates the degree to which the independent variable or variables reflect the variance in the dependent variable. It extends from 0 to 1, with 1 denoting that the model and data are perfectly aligned.

$$R^2 = \frac{\sum_{a=1}^{n}(x_a - x_m)^2 - \sum_{a=1}^{n}(x_a - y_a)^2}{\sum_{a=1}^{n}(x_a - x_m)^2} \tag{3}$$

*4.4. SHAP analysis*

To comprehend how each input characteristic affects the model's output, sensitivity analysis is crucial. Model explanation is especially well suited for machine learning models [28]. Even still, the model's choices are still unknown because it is a "black box." The complex nonlinear behavior of these variables necessitates careful investigation. Because of its explicable behavior, the study used Shapley additives explanations (SHAP) analysis. SHAP estimates the features' contribution using ideas from game theory. SHAP takes into account the marginal contributions while calculating each feature's contribution. Equation (4) provides the information on feature's contribution [40].

$$\phi_i(v,x) = \frac{1}{N!}\sum_{S\subseteq N\setminus x_i}[|S|!\,(N-|S|-1)!][v(S\cup\{x_i\}) - v(S)] \tag{4}$$

Here, v(S) is the model output of a subset of features, and N is the number of features. S. If the SHAP explainable model h(x') possesses the three desired characteristics of consistency, missingness, and local accuracy, it is said to have a unique solution. Equation (4) is used to calculate h(x') [40].

$$h(x') = \phi_0 + \sum_{i=1}^{M} \phi_i x' \tag{5}$$

where, $x' \in \{0,1\}^M$. $x'$ equals 0 if a feature is not observed.

## 5. Results and discussion

Based on Markosyan et al. (1971), Section 2.1 covered the use of cellulose acetate for cigarette filters, including the components and processes used in their manufacture, such as the creation of fibrous materials and the measuring of air resistance. Research by Sadabad et al. (2019) on functional nanocoatings that improved the efficiency of cigarette filters by using natural antioxidants and phenolic compounds obtained from plants was emphasized in Section 2.2. Furthermore, Section 2.3 uses sophisticated computational techniques to assess the ability of single-walled carbon nanotubes (SWCNT) loaded with palladium (Pd) and nickel (Ni) to adsorb tobacco-specific nitrosamines (NNK) from smoke.

**ML modeling results:**

| Model | Training RMSE | Testing RMSE | Training R2 Score | Testing R2 Score |
|---|---|---|---|---|
| **XGBoost** | 0.16 | 0.46 | 1.00 | 1.00 |
| **CatBoost** | 0.16 | 0.23 | 1.00 | 1.00 |
| **Extra Trees** | 11.90 | 11.88 | 0.98 | 0.98 |
| **Random Forest** | 1.61 | 0.68 | 1.00 | 1.00 |

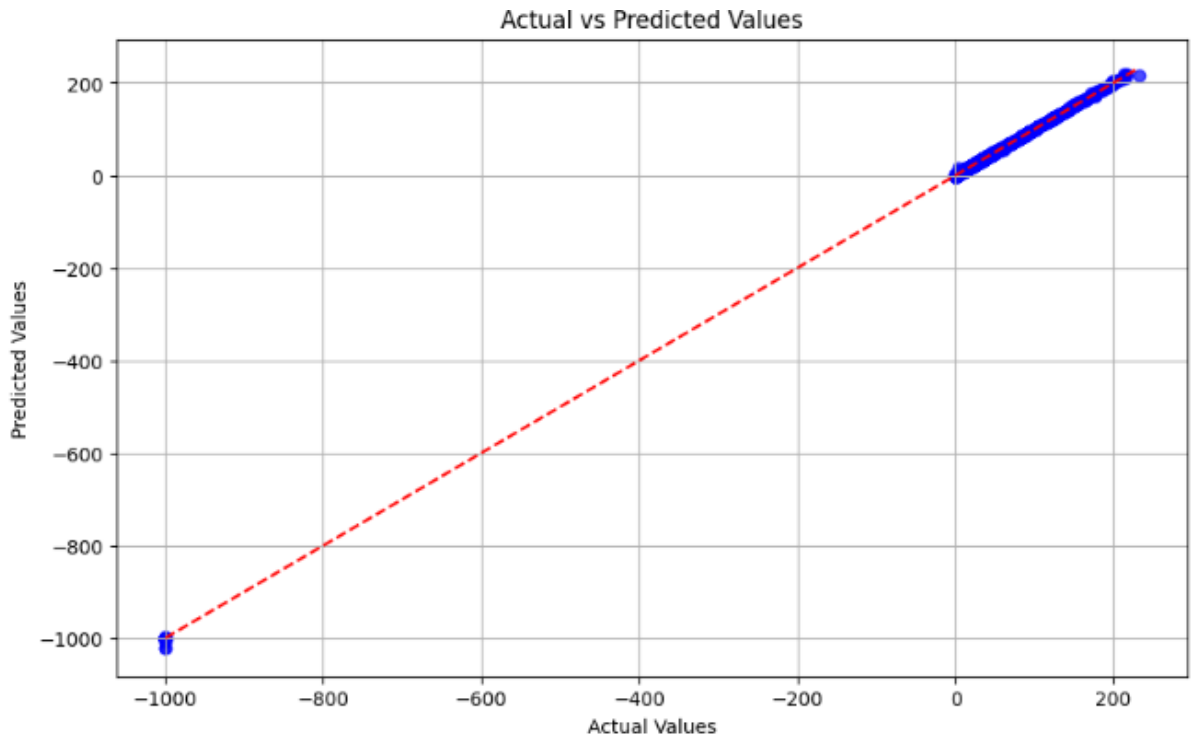

Figure : Actual vs predicted plot for XGBoost model

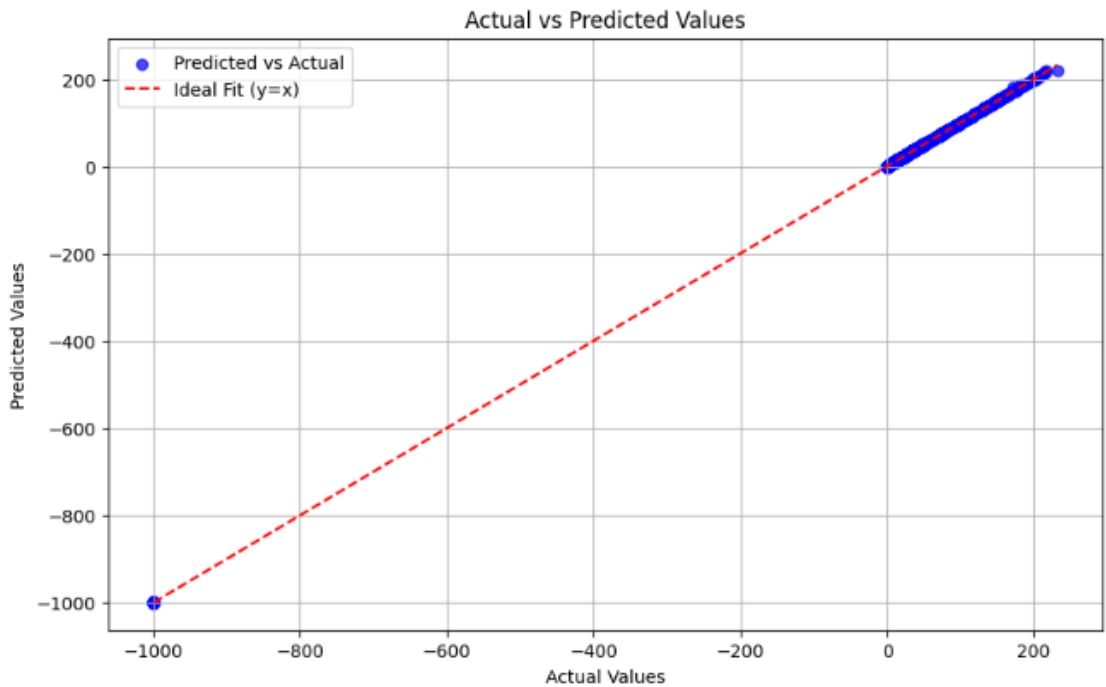

Figure : Actual vs predicted plot for catboost model

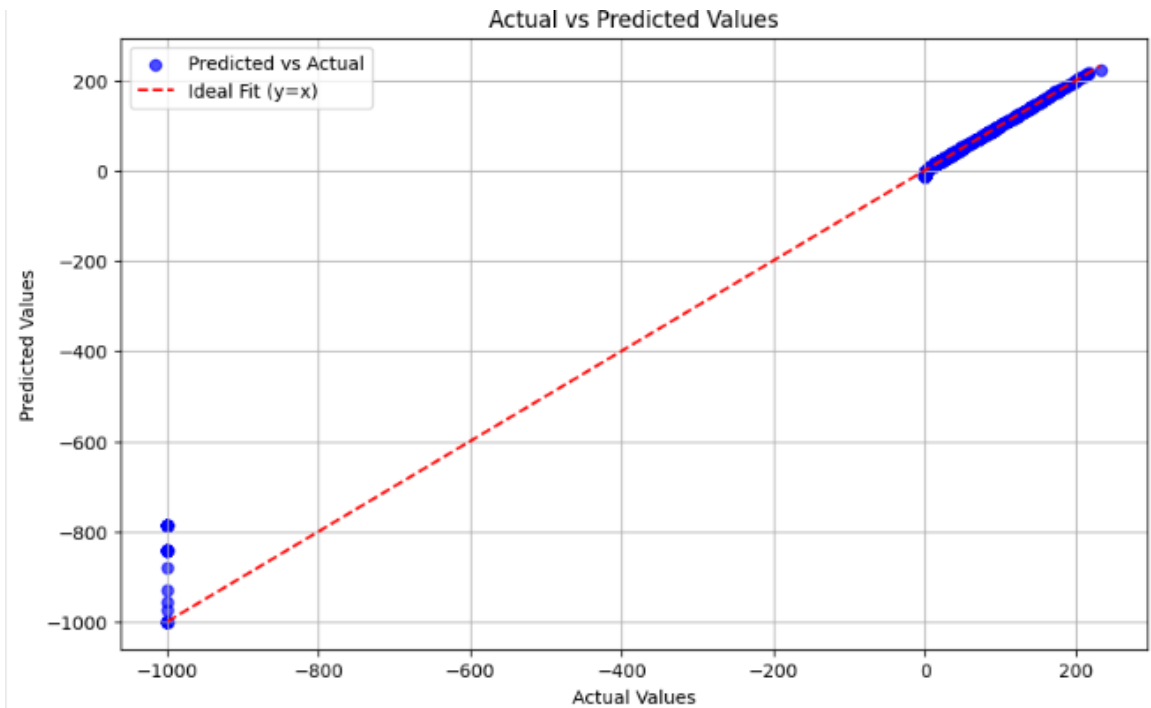

Figure: Actual vs predicted plot for Extra Trees Regressor model

Nested Cross validation results:

| Fold | RMSE |
|---|---|
| 1 | 14.13 |
| 2 | 11.67 |
| 3 | 16.69 |
| 4 | 14.25 |
| 5 | 15.56 |
| 6 | 9.30 |
| 7 | 10.62 |
| 8 | 15.02 |
| 9 | 15.22 |
|  | 16.01 |
| **Mean RMSE** | **13.85** |
| **Standard Deviation of RMSE** | **2.34** |

This table now includes the **Mean RMSE** and **Standard Deviation** values

Table: Models hyperparameters used

| Model | Hyperparameters |
|---|---|

| XGBoost | n_estimators=500, learning_rate=0.1, max_depth=6, subsample=0.8, colsample_bytree=0.8, random_state=42 |
|---|---|
| **CatBoost** | iterations=500, learning_rate=0.1, depth=6, l2_leaf_reg=3, random_seed=42, verbose=0 |
| **Extra Trees** | (Not provided, default parameters used for Extra Trees) |
| **Random Forest** | n_estimators=200, max_depth=10, min_samples_split=5, min_samples_leaf=2, max_features='sqrt', random_state=42 |

**SHAP explainable AI results:**

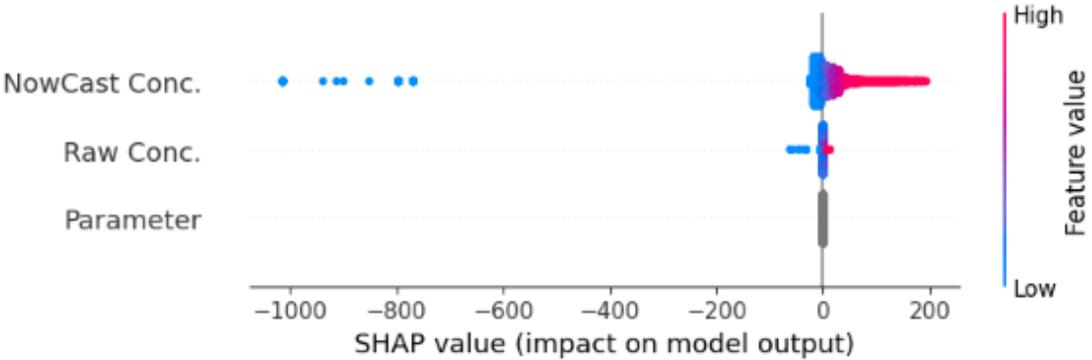

Fig : SHAP plot (impact on model output)

The influence of different characteristics in forecasting the Air Quality Index (AQI) for Kathmandu is shown in this SHAP summary plot. The SHAP values, which show how each feature contributes to the model's output, are represented on the x-axis. When a feature has a positive SHAP value, the anticipated AQI rises; when it has a negative value, the AQI falls. The feature values are reflected in the color gradient, with pink or red denoting high feature values and blue denoting low feature values.

The cluster of red points on the right indicates that the "NowCast Concentration" feature significantly improves AQI forecasts, especially at higher values. This suggests that there is a considerable correlation between elevated AQI and high NowCast Concentration levels. Low NowCast Concentration values (blue), on the other hand, barely affect the forecasts. The "Raw Concentration" feature has a variety of effects; in certain cases, it has a significant impact on both rising and falling AQI, indicating that its contribution varies. Finally, the majority of SHAP values are near zero, suggesting that the "Parameter" feature has no impact on the AQI forecast. All things taken into consideration, this graphic emphasizes how crucial NowCast Concentration is in forecasting the AQI for Kathmandu.

According to the SHAP investigation, NowCast Concentration has the biggest influence on AQI forecasts, which is directly related to the necessity of efficient filtration of these contaminants in order to safeguard public health. By focusing on the main causes of poor air quality that the model indicated, the HA filter can remove more than 98% of PM2.5 and 99.24% of PM10, which is consistent with the SHAP results. This demonstrates a clear correlation: the pollutants that the HA filter successfully reduces are the same pollutants that the SHAP analysis found to be the most significant (PM2.5 and PM10), offering a workable solution to the problems the model's projections predicted.

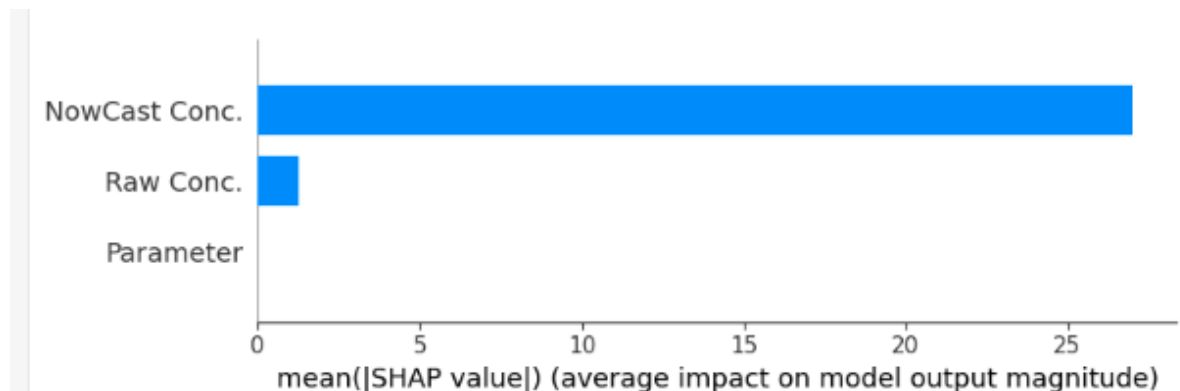
Fig: SHAP feature importance plot

According to a study by de Falco et al. (2020), Section 2.5 examined the effectiveness of phenolic substances, such as gallic acid and epigallocatechin, in lowering harmful carbonyl species in e-cigarettes. Last but not least, Section 2.6 described ways to improve medical mask filters by functionalizing polyphenols, and Table 4 provided an overview of current developments in filter-making technologies, highlighting a number of creative approaches meant to increase filter efficacy and efficiency in various applications.

Recent developments in cigarette filters, as discussed in Section 3.1, have used biological materials, advanced filtration methods, and creative designs to improve user experience, sustainability, and filtering efficiency. Researchers have looked into innovative filter designs that might lessen the environmental effect of cigarette manufacture, such as creating chambers for smoke flow and filtering using vegetable flour paste (Alessandro, 2021). High repeatability and efficacy in filtering particles were demonstrated by sophisticated filtration systems that were created to mimic the deposition of respiratory particles in people (Hao et al., 2022). Additionally, different smoke channel sizes were included into newer filter designs to increase hardness, decrease tar visibility, and boost user satisfaction (Yu & Kenichi, 2014). The extent to which these filters may considerably reduce the health dangers connected with smoking was still up for discussion despite these advancements. The literature studies on the latest developments in cigarette filter technology were included in Table 5.

Medical masks used a variety of filters in Section 3.2 to increase their efficacy against airborne diseases and particulate matter; the composition and structure of these masks had a significant impact on their performance. Advanced masks like N95 obtained filtration efficiency of 97-99% against tiny aerosols, whereas medical masks showed efficiencies ranging from 52% to 77% (Tomkins et al., 2024). High-filtration composites with layers of activated carbon and multilayer masks for improved aerosol filtering were examples of innovative filter designs (Songbo, 2020; Minqiang, 2018). In order to achieve high efficiency and antibacterial qualities, research has looked at other materials, such as bacterial cellulose mixed with natural extracts (Jonsirivilai et al., 2022). Comprehensive literature evaluations of medical filters, as shown in Table 7, emphasized a number of cutting-edge strategies, such as biodegradable and very efficient filters manufactured from new materials that enhanced defense against airborne diseases like SARS-CoV-2.

Modern filters in cigarettes and masks have increased their ability to shield users from dangerous gases and particles, as covered in Section 3.3. Safety, user comfort, and filtering efficiency were all improved by developments in filter materials and designs. Cigarette filters, mask filters, and gas masks and respirators were the three categories into which the development of filtering technologies was divided. New filter-sorbing materials (FSM) that fused glass or polymer fibers with activated carbon were among the recent developments that improved defense against a range of gases and aerosols, including ammonia (Talipova et al., 2023). Small, low-resistance filters also increased ventilation without sacrificing filtering effectiveness. In terms of mask filters, developments include gel breaker agents efficiently eliminated particle matter while maintaining mask permeability, and antibacterial copper wire wrapped around resin fibers inactivated bacteria and viruses (Kodera et al., 2010; Shuhua et al., 2015). New drying techniques for cigarette filters improved moisture content and decreased toxic emissions, making smoking healthier (Antonella & Monzoni, 2018; Shi et al., 2019). Table 8 highlighted the impact of these developments.

The need for appropriate filters in masks and cigarettes was emphasized in Section 3.4 as being essential for public health, particularly in reducing the hazards associated with hazardous substances and airborne illnesses. As shown in Table 9, recent studies have brought to light creative ways to improve filter safety and effectiveness, offering information that may help guide policy decisions in these domains. The use of sodium dihydrogen phosphate to collect and inactivate aerosolized viruses and the incorporation of UV-LEDs to disinfect inhaled and exhaled air were two examples of advancements in mask filters (Lee et al., 2021; Christoph & Chih, 2020; Haitao, 2017). The potential for cigarette filter materials, particularly cellulose acetate, to produce carcinogenic fibers has raised concerns, underscoring the need for greater consumer education and regulatory actions (Hastrup et al., 2001). In order to lessen hazardous exposure, a literature study that is described in Table 10 highlighted the significance of creating antimicrobial filters for masks and proposed that comparable developments be investigated for cigarette filters. Together, the research emphasized the importance of stringent safety evaluations, the use of biodegradable materials, and the requirement for regulatory revisions to guarantee the efficacy and security of filtering systems in both situations.

Section 4 looked at how cigarette and mask filters affected the environment, emphasizing how much of an influence they had on air quality and public health. Filtering membranes were necessary for usage in masks and cigarettes, but their negative environmental effects needed to be taken into account before they were used. With an estimated 7 billion thrown away annually, cigarette filters—which are mostly made of cellulose acetate—were the most littered item in the world (Kadir & Sarani, 2015). These filters put aquatic life and human health at risk by releasing harmful compounds into the environment (Nitschke et al., 2023). Additionally, mask filters showed a significant reduction in filtration capacity when exposed to cigarette smoke, which may increase exposure to dangerous chemicals (Heo et al., 2022). Table 11's literature review highlighted the pollution that cigarettes and face masks produce, describing several research that looked into the goods' effects on the environment. The COVID-19 epidemic increased the manufacturing of disposable face masks, exacerbating waste problems worldwide, particularly in nations like Nepal where misconceptions regarding mask materials still exist. The primary component of masks worn in Nepal's Kathmandu Valley, according to research by Kattel et al. (2023), is polypropylene microplastic, which may have negative environmental effects. Furthermore, contradicting popular belief, 2017 research by John and Ross polled 1,204 cigarette dealers in Nepal, gathering 4,307 empty packs and discovering that domestic sales, not illegal imports, are the main source of low-quality cigarettes. Cigarettes of poor quality are undoubtedly less effective at preventing carcinogenic substances.

It became clear that a comprehensive approach was required to address the negative effects that single-use face masks and cigarette waste were having on the environment and human health. To reduce soil and water contamination, stricter laws governing the biodegradability and disposal of cigarette filters were required, in addition to public education initiatives encouraging appropriate disposal methods. Furthermore, the creation of eco-friendly filter substitutes may aid in lowering pollution from dangerous chemicals and polymers. According to studies by Beutel et al. (2021) and De-la-Torre et al. (2023), in order to reduce microplastic waste and leachate formation, mask pollution has to be addressed by developing reusable or single-use masks composed of biodegradable materials. Education and incentives for appropriate mask disposal might significantly minimize plastic pollution in natural habitats. To address these environmental issues, international cooperation on rules to efficiently manage waste disposal and encourage sustainable material use in product manufacture was crucial.

Markosyan et al. (1971) conducted research on the usage of cellulose acetate fibers in cigarette filters and discovered that the weight of the fibers rose with their diameter. According to this study, the sorption of different tobacco smoke components was made possible by hydrogen bond formation, which increased the quantity of hydroxyl groups. It did not, however, explain how sorption effectiveness was increased for certain harmful chemicals, such 3,4-benzpyrene. The results also demonstrated that the internal fiber surface area increased as the amount of acetate substituted in the fibers decreased.

Additionally, functional nanocoatings modeled after natural polyphenols were studied by Sadabad et al. (2019). Polyphenol solutions oxidized and polymerized when exposed to UV light, forming plant-based polyphenol micropatterns on surfaces. More consistent polyphenolic nanocoatings were made possible by this regulated process, which was supported by antioxidants. The idea of substrate-independent coating precursors is derived from this technique, which takes use of the potent adhesion qualities of mussel foot proteins (MFP) and the capacity of plant phenols to connect with a variety of substrates. But applying these nanocoatings to cigarette filters is a novel concept that needs more investigation to reach its full potential. Even if these coatings have several uses, it might be difficult to guarantee compatibility and adherence between layers as well as between the coating and substrate.

Furthermore, cutting-edge research on functionalized carbon nanotube-based filters by Yoosefian (2018) indicates encouraging potential to lower tobacco-specific nitrosamines in cigarettes. According to calculations, SWCNTs loaded with transition metals have a strong attraction for NNK molecules, although their adsorption energies for complexes combining NNK molecules and Pd/SWCNTs varied somewhat. According to the computed results, these exothermic adsorption energies point to a robust chemisorption interaction between Pd and NNK active sites (Table 13).

**Table 13** The calculated values of the adsorption energies ($E_{ads}$ in eV), total energy ($E_{HF}$ in Hartree) and Dipole moments (DM in debye) of investigated compounds (Yoosefian, 2018).

| Compound | $E_{HF}$ | $E_{ads}$ (adsorption energy) | DM |
|---|---|---|---|
| Complex 1 | -8284.538 | -17.960 | 11.570 |
| Complex 2 | -82824.537 | -17.917 | 16.353 |

When cigarette including the filter made up of such microemulsions was combusted, the hydroxytyrosol microcapsule got heated, absorbed the exposure such as oral cavity that could affect eye or lung organ of the consumer. The research work conducted to reduce the toxic carbonyl e-cigarettes has presented a good method to reduce toxicity in cigarettes (de Falco et al., 2020). The work showed that the addition of gallic acid, hydroxytyrosol and epigallocatechin gallate could reduce the levels of aerosols of vaped e-cigarettes, methylglyoxal and glyoxal. It was highlighted by the liquid chromatography mass spectrometry analysis that the formation of covalent adducts takes place between the aromatic rings and di-carbonyl in both e-liquids and vaped samples which showed that the dicarbonyls formed in the e-liquids were formed as degradation products of propylene glycol and glycerol and vaping and the short term cytotoxic analysis on two lung cellular models showed that the dicarbonyl-polyphenol were not cytotoxic even though carbonyl did not improve cell viability.

Lastly, development of wipes and filters functionalized with polyphenols (Fillat et al., 2012), showed a modern and more effective method to build a filter that can even be utilized in commercial cigarettes too. The spectra of cellulose before and after treatments were baseline corrected and normalized at 1058 $cm^{-1}$, the major absorbance peak that reflected the carbohydrate backbone of cellulose. The main spectral characteristics of cellulose were in the 900–1700 $cm^{-1}$ range. The effect that laccase-catechin treatment had on cellulose was particularly striking in the (1200–1600 $cm^{-1}$) that corresponds to the range of catechin. For both spectra of multi-substituted aromatic rings and to the vibration of the hydroxyl groups substituted on aromatic rings, same peak of vibrations of valence were observed around 1035 $cm^{-1}$, 1058 $cm^{-1}$, 1100 $cm^{-1}$ and 1162 $cm^{-1}$. No new peaks were noted between 1060 $cm^{-1}$ and 1150 $cm^{-1}$ that corresponded to the peaks of vibration of ether groups due to the grafting phenomenon that occurred between hydroxyl groups of cellulose and catechin. The reduction in the intensities of the peaks at 1058 $cm^{-1}$ and at 1035 $cm^{-1}$ was seen which was probably due to the substation of alcohol function on cellulose. A conclusion could be made from data obtained that changed in spectra of the modified cellulose wipe proved the efficiency of treatment process.

The higher sorption of 3,4-benzpyrene and related compounds was not explained by Markosyan et al. (1971), despite a review of important investigations and methodology. Additionally, studies conducted in the last several decades have shown that cellulose acetate filters do not completely reduce the harmful compounds produced during cigarette combustion (Pauly et al., 2002; Harris, 2011; Joly & Coulis, 2018; McKee et al., 1978; Evans et al., 2022). On the other hand, Yoosefian (2018) suggested a productive technique and found that SWCNTs loaded with transition metals had a high affinity for NNK molecules. But before being used commercially, the use of SWCNTs in cigarette filters is still a matter of worry that needs more research. Furthermore, there was no investigation into the actual uses of SWCNTs; instead, the study concentrated mostly on their affinity for NNK. According to Catel-Ferreira et al. (2015), two layers of Kimwipes® Lite (from Kimberly-Clark®) treated with PEI solutions at a concentration of 4.4% w/v for 15 minutes produced the most effective virus capture factor (f), with a f value of $3 \times 10^5$. However, further experimental research is needed to confirm these results.

The biodegradable face mask filter's efficacy and appropriateness for usage in airborne particulate matter conditions, including cigarette smoke and other contaminants discussed in (Choi et al., 2021), were proved by the trials that were done on it. The created filter had remarkable particulate matter (PM) removal efficiency, up to 98.3% for PM2.5 and 99.24% for PM10, which were equivalent to those of commercial N95 masks, according to the data. The cumulative dose of particulate matter inhaled by employees in one day is given by equation (3) (Chang et al., 2023).

$$\text{Dose of the particulate on lung} = \sum_{i=1}^{t}(C \times V_e \times 10^{-3}) \tag{3}$$

Where:
- C is the average particulate matter concentration measured within a minute time interval.
- t is the work time in 1 day in minutes.

- $V_e$ is the inhalation rate defined as the volume of air that enters the lungs per minute.

For particulate matter exposure within the range 1 µg/m³ to 100 µg/m³ exposed to human body (C) and t be varied from 1 to 600 minutes as a time period of exposure (10 hours) within a day. Dose for the normal inhalation rate ($V_e$) within normal range of 12 to 18 breaths/minute (Cleveland Clinic, 2023) is represented by Fig. 9 for PM2.5 and PM10 concentration of 98.3% and 99.245% of C respectively to find the toxic effect prevented by filter on human body using equation (3).

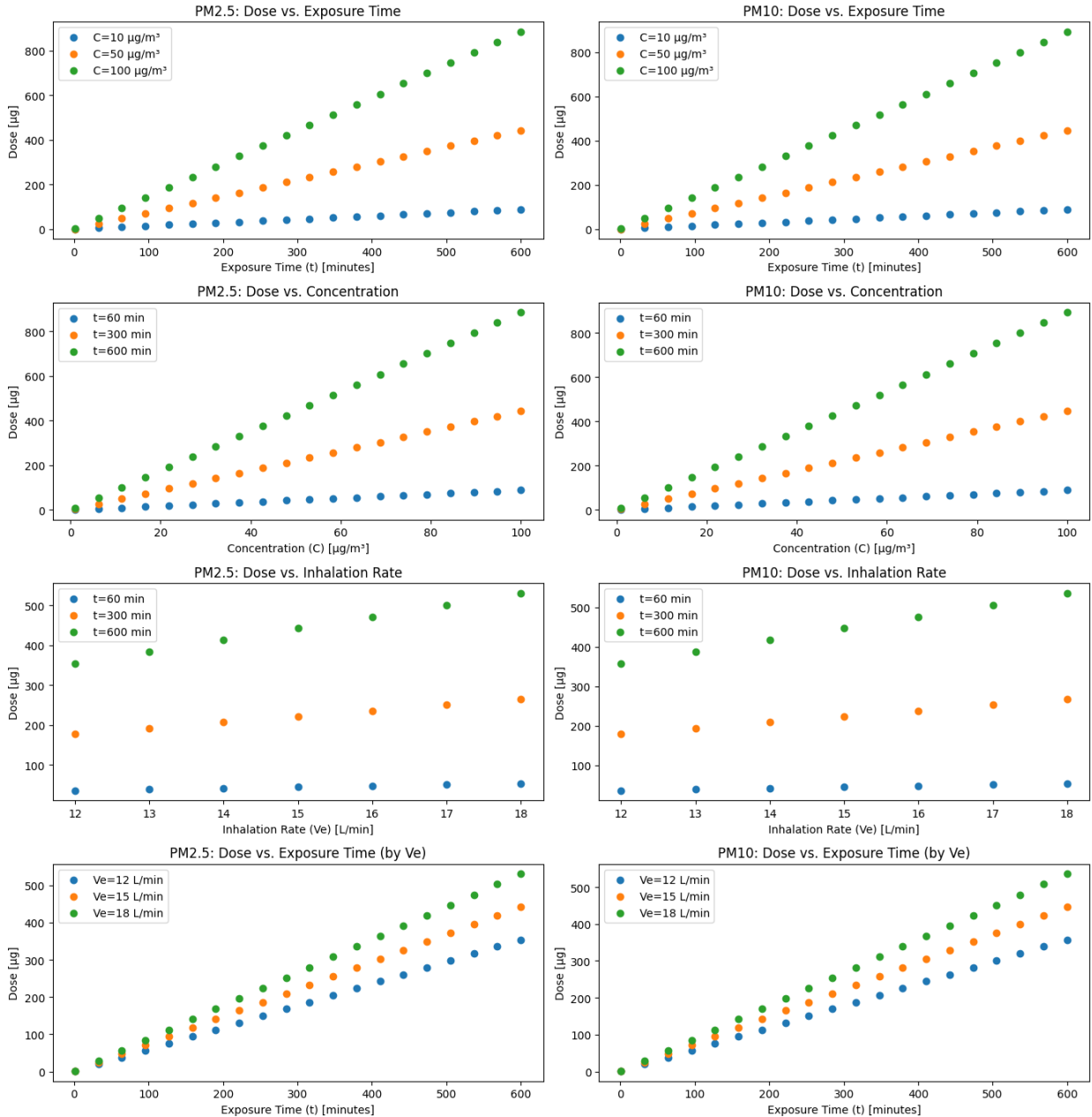

Fig. 9. Scatter plots to show effect of exposure of PM2.5 and PM10 concentration of 98.3% and 99.245% of C at different time periods of exposure and inhalation rate.

Simialrly, Plotly plot in Python were obtained to determine the dosage on human lungs by the variation of parameters within particulate matter exposure within the range 1 μg/m$^3$ to 100 μg/m$^3$ exposed to human body (C) and t be varied from 1 to 600 minutes as a time period of exposure (10 hours) within a day, Dose for the normal inhalation rate ($V_e$) within normal range of 12 to 18 breathes/minute is represented by Fig. 10 for PM2.5 and PM10 concentration of 98.3% and 99.245% of C respectively to find the toxic effect prevented by filter on human body using equation (3).

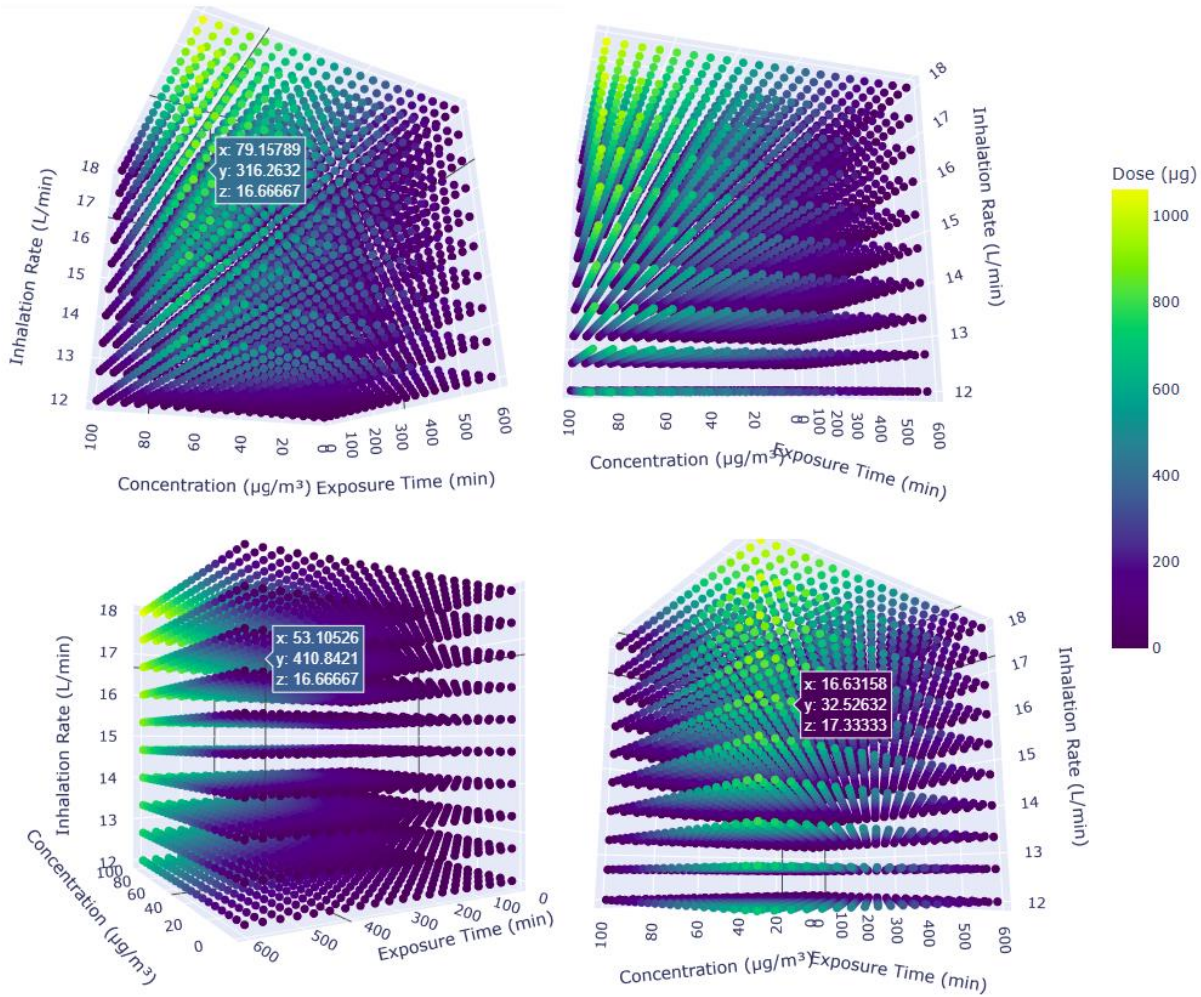

Fig. 10. Dosage on body as a result of varying parameters.

The Fig. 9 and Fig. 10 show that the biodegradable face mask filter's efficacy on preventing the PM2.5 and PM10 materials due to their removal efficacy as discussed previously and appropriateness for usage in airborne particulate matter conditions, including cigarette smoke and other contaminants discussed in (Choi et al., 2021), were proved by the trials that were done on it. The created filter had remarkable particulate matter (PM) removal efficiency, up to 98.3% for PM2.5 and 99.24% for PM10, which were equivalent to those of commercial N95 masks and this implies that tiny particulate matter, including dangerous substances included in cigarette smoke, may be efficiently captured by the filter. Furthermore, the filter maintained a low pressure drop of only 59 Pa, which is essential for extended usage of face masks since it ensures user comfort when breathing. Both Fig. 9 and Fig. 10 are based on application of equation (3). Both of them show how the efficiency of the filter explained by Choi et al. (2020), can have high efficacy in preventing two of the most toxic materials PM2.5 and PM10. Simply, we can consider them as representation of prevention of dosage of particulates due to use of the same filter explained in their research (Choi et al., 2020). The plots show that the filter has a very high efficiency and efficacy to prevent the smokers or inhalers from toxic smokes and vapors consisting harmful particulates.

The biodegradable filter was very efficient in humid situations, as during exhalation or in rainy weather, because it showed minimal performance loss even when wet, in contrast to conventional filters that lose effectiveness when exposed to moisture. In addition, the filter completely biodegraded in four weeks in composting soil, demonstrating its advantages for the environment over conventional single-use masks that greatly increase landfill trash. Overall, the results obtained from above discussions can be summarized in Table 14.

**Table 14** Results of comparison of different filters.

| Feature | Reference | Function | Results | Limitations |
|---|---|---|---|---|
| High Filtration Efficiency | (Choi et al., 2021) | Achieves removal of airborne particles | Over 98% for PM2.5 and 99.24% for PM10 | Efficiency for other particle types unknown |
| Low Pressure Drop | (Choi et al., 2021) | Enhances user comfort by maintaining low pressure drop | Pressure drop: 59 Pa, allowing easier breathing compared to conventional filters | - |
| Moisture Resistance | (Choi et al., 2021) | Ensures consistent filtration effectiveness in humid conditions | Negligible performance loss when exposed to moisture | - |
| Biodegradability | (Choi et al., 2021) | Addresses environmental concerns through material decomposition | Full decomposition within four weeks (composting soil) | Biodegradation rate and impact need further study |
| Permanent Electrostatic Charge | (Choi et al., 2021) | Enhances PM adsorption with permanent ionic charges | Maintains effectiveness over time, even in humid conditions, unlike traditional filters | - |
| Multi-Usability | (Choi et al., 2021) | Supports reuse without significant performance loss | Allows multiple uses | - |
| Comfortable Breathing Environment | (Choi et al., 2021) | Combines features to ensure a pleasant | Suitable for extended wear in various settings | - |

| | | breathing experience | | |
|---|---|---|---|---|
| Versatility | (Choi et al., 2021) | Adaptable for use in different environments | Effective in urban areas, smoke exposure, or high particulate settings | - |
| Limitations of Cellulose Acetate Filters | (Markosyan et al., 1971; Pauly et al., 2002; Harris, 2011; Joly and Coulis, 2018; McKee et al., 1978; Evans et al., 2022) | Highlights gaps in existing filtration technologies | Ineffective against toxic chemicals in cigarette smoke | Requires alternative filtration solutions |
| Transition Metal-Loaded SWCNTs | (Yoosefian, 2018) | Investigates advanced materials for cigarette smoke filtration | Shows promise for NNK capture | Needs further research for practical application |
| Virus Capture Factor | (Catel-Ferreira et al., 2015) | Captures viruses with functionalized materials | Best factor with PEI solutions (f = $3 \times 10^5$) | Biodegrability issues |
| Cigarette Pollution Concerns | (Guo et al., 2023; Beutel et al., 2021) | Addresses environmental pollution from cigarette waste | Highlights need for effective filtration | Filtration efficiency is low to achiece high filtration |
| Environmental Effects | (Lal & Swaroop, 2017; Amuah et al., 2022) | Examines impact of waste on environment | Significant contribution to plastic pollution | Emphasizes sustainable filtering solutions |

In summary, the biodegradable cigarette filter addressed significant environmental issues related to disposable masks while effectively trapping particulate matter. Because of its high filtration effectiveness, low pressure drop and moisture resistance, it may be used in a variety of situations, especially those with high air pollution levels, such as cities and locations where people are exposed to a lot of cigarette smoke. The filter is particularly suitable for areas where cigarette smoke is present because to its effective collection of tiny particulate matter, which helps to lower hazardous exposure for both smokers and non-smokers. This biodegradable filter offers a viable solution for individual protection against airborne pollutants while also supporting environmental sustainability, which is important given the continuous need for efficient face masks brought on by the COVID-19 pandemic and air quality issues. Its ability to effectively filter cigarette smoke highlights its potential for wider use in public safety and health including in face masks. By tackling environmental and public health issues, the biodegradable face mask filter covered by Choi et al. (2021) can have a major positive impact on poor nations like Nepal. PM2.5 and PM10, two dangerous airborne particles that are common in cities with poor air quality, are successfully captured by its high filtration effectiveness. Such filters can be effective to remove toxic ingredient like nitrosamines during smoking process as well as toxic compounds in air when used in face masks. The little pressure-drop guarantees user comfort and promotes extended usage, especially in humid environments where conventional masks might not work as well. Additionally, the eco-friendly and biodegradable nature of the filter minimizes plastic waste, which is in line with sustainable methods that are essential for Nepal's waste management issues.

Various good features like its its longevity and versatility make it cost-effective, and the possibility of local manufacture might stimulate economic expansion. The use of such cutting-edge filters can also raise community health awareness, encouraging healthier lifestyle choices and creating a healthier atmosphere. All things considered, the application of this filter may be extremely important for enhancing sustainability and public health in Nepal difficulties.

**6. Conclusion**
In conclusion, the 98-99.5% effective biodegradable filter appears to be a ground-breaking way to enhance public health in Nepal, especially when it comes to reducing the negative impacts of airborne pollutants and cigarette smoke. Its sophisticated design meets the demands of metropolitan populations exposed to poor air quality by guaranteeing a minimal pressure drop and moisture resistance, making it incredibly pleasant for extended usage. This filter is a great option for both cigarette filters and medical masks in underdeveloped nations like Nepal because it combines high filtering performance with environmental sustainability, protecting users from harmful particulate matter while also helping to reduce plastic waste.

The machine learning and SHAP results help identify the key pollutants that affect AQI, and the HA filter offers a sustainable way to deal with these pollutants. This highlights how crucial it is to combine data-driven insights with technological solutions to enhance air quality and health outcomes in places like Nepal.

Jiang, Y., Zhou, C., Khan, A., Zhang, X., Tursunay Mamtimin, Fan, J., Hou, X., Liu, P., Han, H., & Li, X. (2024). Environmental Risks of Mask Wastes Binding Pollutants: Phytotoxicity, Microbial Community, Nitrogen and Carbon Cycles. Journal of Hazardous Materials, 476, 135058–135058. https://doi.org/10.1016/j.jhazmat.2024.135058

John, R. M., & Ross, H. (2017). Illicit cigarette sales in Indian cities: findings from a retail survey. Tobacco Control, 27(6), 684–688. https://doi.org/10.1136/tobaccocontrol-2017-053999

Joly, F. X., & Coulis, M. (2018). Comparison of cellulose vs. plastic cigarette filter decomposition under distinct disposal environments. *Waste Management*, *72*, 349-353.

Jonsirivilai, B., Torgbo, S., & Sukyai, P. (2022). Multifunctional filter membrane for face mask using bacterial cellulose for highly efficient particulate matter removal. *Cellulose*, *29*(11), 6205–6218. https://doi.org/10.1007/s10570-022-04641-3

Ju, J. T. J., Boisvert, L. N., & Zuo, Y. Y. (2021). Face masks against COVID-19: Standards, efficacy, testing and decontamination methods. Advances in Colloid and Interface Science, 292, 102435. https://doi.org/10.1016/j.cis.2021.102435

Kadir, A. A., & Sarani, N. A. (2015). Cigarette Butts Pollution and Environmental Impact – A Review. Applied Mechanics and Materials, 773-774(1662-7482), 1106–1110. https://doi.org/10.4028/www.scientific.net/amm.773-774.1106

Khanal, G., Karna, A., Kandel, S., Sharma, H. K., & Ward, K. (2023). Prevalence, Correlates, and Perception of E-cigarettes among Undergraduate Students of Kathmandu Metropolitan City, Nepal: A Cross-Sectional Study. *Journal of Smoking Cessation*, *2023*, e11. https://doi.org/10.1155/2023/1330946

Kim, Jong, Yeol., Kim, Soo, Ho., Rhee, Moon, Soo., Park, Jin, Won., Kim, Yonug, Hoi. (2009). 10. Cigarette filters and cigarettes, comprising paper filters treated with flavoring natural botanical extracts.

Konstantinou, E., Fotopoulou, F., Drosos, A., Dimakopoulou, N., Zagoriti, Z., Niarchos, A., ... & Poulas, K. (2018). Tobacco-specific nitrosamines: A literature review. *Food and chemical toxicology*, *118*, 198-203.

Konstantinou, E., Fotopoulou, F., Drosos, A., Dimakopoulou, N., Zagoriti, Z., Niarchos, A., Makrynioti, D., Kouretas, D., Farsalinos, K., Lagoumintzis, G., & Poulas, K. (2018). Tobacco-specific nitrosamines: A literature review. *Food and Chemical Toxicology*, *118*, 198–203. https://doi.org/10.1016/j.fct.2018.05.008

Koopmans, T. (1934). Über die Zuordnung von Wellenfunktionen und Eigenwerten zu den einzelnen Elektronen eines Atoms. *physica*, *1*(1-6), 104-113.

Kozlowski, L. T., & O'Connor, R. J. (2000). Official cigarette tar tests are misleading: use a two-stage, compensating test. *The Lancet*, *355*(9221), 2159-2161.\\
50

Magee, P. N. (1989). The experimental basis for the role of nitroso compounds in human cancer. Cancer surveys, 8(2), 207-239.

Magee, P. N. (1996). Nitrosamines and human cancer: introduction and overview. European Journal of Cancer Prevention, 5, 7-10.

Markosyan, D. E., Pirverdyan, A. I., Mokhnachev, I. G., & Perepechkin, L. P. (1971). Cellulose acetate fibre for cigarette filters. *Fibre Chemistry*, *2*(3), 292-293.

Martí, M., Tuñón-Molina, A., Aachmann, F., Muramoto, Y., Noda, T., Takayama, K., & Serrano-Aroca, Á. (2021). Protective Face Mask Filter Capable of Inactivating SARS-CoV-2, and Methicillin-Resistant Staphylococcus aureus and Staphylococcus epidermidis. *Polymers*, *13*(2), 207. https://doi.org/10.3390/polym13020207

Martí, M., Tuñón-Molina, A., Aachmann, F., Muramoto, Y., Noda, T., Takayama, K., & Serrano-Aroca, Á. (2021b). Protective Face Mask Filter Capable of Inactivating SARS-CoV-2, and Methicillin-Resistant Staphylococcus aureus and Staphylococcus epidermidis. *Polymers*, *13*(2), 207. https://doi.org/10.3390/polym13020207

Martí, M., Tuñón-Molina, A., Finn Lillelund Aachmann, Muramoto, Y., Noda, T., Takayama, K., & Ángel Serrano-Aroca. (2020). Protective face mask filter capable of inactivating SARS-CoV-2, and methicillin-resistant Staphylococcus aureus and Staphylococcus epidermidis. BioRxiv (Cold Spring Harbor Laboratory). https://doi.org/10.1101/2020.11.24.396028

Martínez, L., Ros, G., & Nieto, G. (2018). Fe, Zn and Se bioavailability in chicken meat emulsions enriched with minerals, hydroxytyrosol and extra virgin olive oil as measured by Caco-2 cell model. *Nutrients*, *10*(8), 969.

Martínez, L., Ros, G., & Nieto, G. (2018). Hydroxytyrosol: Health benefits and use as functional ingredient in meat. Medicines, 5(1), 13.

Matt, G. E., Quintana, P. J., Destaillats, H., Gundel, L. A., Sleiman, M., Singer, B. C., ... & Hovell, M. F. (2011). Thirdhand tobacco smoke: emerging evidence and arguments for a multidisciplinary research agenda. *Environmental health perspectives*, *119*(9), 1218-1226.

Matteucci, M., & Napolitano, R. S. (2015). Apparatus and method for making filters.

McGaw, S. B., Jr. (2000). Filter manufacturing process using a combined welding and cutting step.

McKee, J. L., Bohlander, P. J., & Bowermaster, J. (1978). Quantitative analysis of triacetin on cellulose acetate cigarette filters. *Tob. Sci*, *22*, 16-18.

McNeill, A., Etter, J. F., Farsalinos, K., Hajek, P., Le Houezec, J., & McRobbie, H. (2014). A critique of a World Health Organization-commissioned report and associated paper on electronic cigarettes. Addiction (Abingdon, England), 109(12), 2128-2134.

Medvedev, P., & Karapetjan, A. (2015). Method of production of filter material.

Mi, L., Dai, B., Qin, Y., Zhang, W., Xiong, Z., Wang, Y., & Zhu, T. (2015). Cluster Analysis of Polyphenols and Organic Acids in 11 Different Brand Cigarette Samples at Home and Abroad. *Agricultural Science & Technology*, *16*(10), 2194.

Moscatelli, Romano. (2010). A filtering mask.

Murphy, J., Gaca, M., Lowe, F., Minet, E., Breheny, D., Prasad, K., ... & Proctor, C. (2017). Assessing modified risk tobacco and nicotine products: description of the scientific framework and assessment of a closed modular electronic cigarette. *Regulatory Toxicology and Pharmacology*, *90*, 342-357.

Nadler, J., Velasco, J., & Horton, R. (1983). Cigarette smoking inhibits prostacyclin formation. *The Lancet*, *321*(8336), 1248-1250

Nazarparvar, E., Zahedi, M., & Klein, E. (2012). Density functional theory (B3LYP) study of substituent effects on O–H bond dissociation enthalpies of trans-resveratrol derivatives and the role of intramolecular hydrogen bonds. *The Journal of Organic Chemistry*, *77*(22), 10093-10104.

Nepal | Global Action to End Smoking. (2021, December 21). Global Action to End Smoking. https://globalactiontoendsmoking.org/research/tobacco-around-the-world/nepal/

Neupane, B. B., Mainali, S., Sharma, A., & Giri, B. (2019). Optical microscopic study of surface morphology and filtering efficiency of face masks. *PeerJ*, *7*, e7142. https://doi.org/10.7717/peerj.7142

Ninomiya, Yu., Itabashi, Kenichi. (2014). Cigarette with filter.

Nitschke, T., Bour, A., Bergquist, M., Blanchard, M., Molinari, F., & Almroth, B. C. (2023). Smokers' behaviour and the toxicity of cigarette filters to aquatic life: a multidisciplinary study. Microplastics and Nanoplastics, 3(1). https://doi.org/10.1186/s43591-022-00050-2

O'Dowd, K., Nair, K. M., Forouzandeh, P., Mathew, S., Grant, J., Moran, R., Bartlett, J., Bird, J., & Pillai, S. C. (2020). Face Masks and Respirators in the Fight Against the COVID-19 Pandemic: A Review of Current Materials, Advances and Future Perspectives. Materials, 13(15), 3363. https://doi.org/10.3390/ma13153363

Pandit, P., Maity, S., Singha, K., Annu, Uzun, M., Shekh, M., & Ahmed, S. (2021). Potential biodegradable face mask to counter environmental impact of Covid-19. *Cleaner Engineering and Technology*, *4*, 100218. https://doi.org/10.1016/j.clet.2021.100218

Parveen, N., Singh, H., Vanapalli, K. R., & Goel, S. (2024). Leaching of organic matter from cigarette butt filters as a potential disinfection by-products precursor. *Journal of Hazardous Materials*, *476*, 134976. https://doi.org/10.1016/j.jhazmat.2024.134976

P. Kattel, R. Chalise, Adhikari, A., & R. Khanal. (2023). Pollution Threat by Face Mask after COVID-19 in Nepal. Journal of Nepal Physical Society, 9(1), 116–121. https://doi.org/10.3126/jnphyssoc.v9i1.57745
54

Tricker, A. R., Spiegelhalder, B., & Preussmann, R. (1989). Environmental exposure to preformed nitroso compounds. *Cancer surveys*, *8*(2), 251-

Tueshaus, R., & McGrenera, P. W. (2006). Tubular filter material assemblies and methods.

Using behaviour change interventions to decrease tobacco use in Nepal POLICY BRIEF Using behaviour change interventions to decrease tobacco use in Nepal. (n.d.). Retrieved October 30, 2024, from https://assets.publishing.service.gov.uk/media/59677a0a40f0b60a4000017c/Using-behaviour-change-interventions-to-decrease-tobacco-use-in-Nepal-August-2014.pdf

Verstraete, M., & Proost, G. (2021). Filter elements and methods of manufacturing filter elements.

Viceroy (cigarette). Retrieved from Wikipedia. on 02/01/2023 at

Visioli, F., Galli, C., Plasmati, E., Viappiani, S., Hernandez, A., Colombo, C., & Sala, A. (2000). Olive Phenol Hydroxytyrosol Prevents Passive Smoking–Induced Oxidative Stress. *Circulation*, *102*(18), 2169–2171. https://doi.org/10.1161/01.CIR.102.18.2169

Wei-Hong, Zhong., Hamid, Souzandeh., Yu, Wang. (2018). Stabilized protein fiber air filter materials and methods.

Vitti Gambim, V., Laufer-Amorim, R., Fonseca Alves, R. H., Grieco, V., & Fonseca-Alves, C. E. (2020). A comparative meta-analysis and in silico analysis of differentially expressed genes and proteins in canine and human bladder cancer. *Frontiers in veterinary science*, *7*, 558978.

Wang, D., Williams, B. A., Ferruzzi, M. G., & D'Arcy, B. R. (2013). Microbial metabolites, but not other phenolics derived from grape seed phenolic extract, are transported through differentiated Caco-2 cell monolayers. *Food chemistry*, *138*(2-3), 1564-1573.

Wang, L., Gao, Y., Xiong, J., Shao, W., Cui, C., Sun, N., Zhang, Y., Chang, S., Han, P., Liu, F., & He, J. (2022). Biodegradable and high-performance multiscale structured nanofiber membrane as mask filter media via poly(lactic acid) electrospinning. Journal of Colloid and Interface Science, 606, 961–970. https://doi.org/10.1016/j.jcis.2021.08.079

Wang, L., Li, S., Ahmad, I. M., Zhang, G., Sun, Y., Wang, Y., Sun, C., Jiang, C., Cui, P., & Li, D. (2023). Global face mask pollution: threats to the environment and wildlife, and potential solutions. Science of the Total Environment, 887, 164055–164055. https://doi.org/10.1016/j.scitotenv.2023.164055

Watts, N., Amann, M., Arnell, N., Ayeb-Karlsson, S., Belesova, K., Berry, H., Bouley, T., Boykoff, M., Byass, P., Cai, W., Campbell-Lendrum, D., Chambers, J., Daly, M., Dasandi, N., Davies, M., Depoux, A., Dominguez-Salas, P., Drummond, P., Ebi, K. L., & Ekins, P. (2018). The 2018 report of the Lancet Countdown on health and climate change: shaping the health of nations for centuries to come. *The Lancet*, *392*(10163), 2479–2514. https://doi.org/10.1016/s0140-6736(18)32594-7